\pdfoutput=1

\documentclass[11pt]{article}

\usepackage[final]{acl}

\usepackage{times}
\usepackage{latexsym}

\usepackage[T1]{fontenc}

\usepackage[utf8]{inputenc}

\usepackage{microtype}

\usepackage{inconsolata}

\usepackage{graphicx}

\usepackage{amsmath,amssymb,mathtools}
\usepackage{booktabs}
\usepackage{multirow}
\usepackage{colortbl} 
\usepackage{xcolor}

\title{Does Time Have Its Place? \\
Temporal Heads: Where Language Models Recall Time-specific Information}

\author{Yein Park$^1$, Chanwoong Yoon$^1$, Jungwoo Park$^{1,3}$, Minbyul Jeong$^{2}$\thanks{Corresponding authors}, Jaewoo Kang$^{1,3}$\footnotemark[1] \\
  $^1$Korea University \quad
  $^2$Upstage AI \quad
  $^3$AIGEN Sciences \\
  \{522yein, cwyoon99, jungwoo-park, kangj\}@korea.ac.kr
}

\begin{document}
\maketitle
\begin{abstract}
While the ability of language models to elicit facts has been widely investigated, how they handle \emph{temporally changing} facts remains underexplored.
We discover \textbf{Temporal Heads}, specific attention heads that primarily handle temporal knowledge, through circuit analysis.
We confirm that these heads are present across multiple models, though their specific locations may vary, and their responses differ depending on the type of knowledge and its corresponding years.
Disabling these heads degrades the model's ability to recall time-specific knowledge while maintaining its general capabilities without compromising time-invariant and question-answering performances.
Moreover, the heads are activated not only numeric conditions (\emph{“In 2004”}) but also textual aliases (\emph{“In the year ...”}), indicating that they encode a temporal dimension beyond simple numerical representation.
Furthermore, we expand the potential of our findings by demonstrating how temporal knowledge can be edited by adjusting the values of these heads\footnote{Our datasets and code are publicly available at \href{https://github.com/dmis-lab/TemporalHead}{https://github.com/dmis-lab/TemporalHead}}.
\end{abstract}

\section{Introduction}
\begin{quote}
    \emph{``Remembrance of things past is not necessarily the remembrance of things as they were.''~\citep{proust}}
\end{quote}
This profound and intricate relationship between memory and truth resonates deeply with one of the central challenges in modern artificial intelligence. 
While large language models (LLMs) like GPTs~\citep{chatgpt, gpt4omini, gpto1} and LLaMA families~\citep{llama, llama2, llama3} have demonstrated remarkable capabilities in leveraging factual knowledge, they face a unique challenge that mirrors human memory: 
the accurate representation of \emph{temporal knowledge}—facts that transform across different time points.

Unlike static facts (e.g., “The capital of France is Paris”), many real-world facts change over time (e.g., a politician’s term in office, a sports player’s team membership in a given year). 
This time-evolving nature necessitates that LLMs accurately capture such change.
To do so, they must not only track newly updated facts within a specific timeline, but also retain historical information across different time periods~\citep{temporalwiki}.
This presents a significant challenge, as models must contend with tracking and reasoning over temporal changes in knowledge~\citep{realtime}.
However, beyond prompting~\citep{serac, chroknowledge} or retrieval-augmentated generation~\citep{rag, hipporag}, the internal mechanisms by which models adapt to temporally evolving facts remain relatively underexplored.

\begin{figure}[t]
\begin{center}
    \includegraphics[width=\columnwidth]{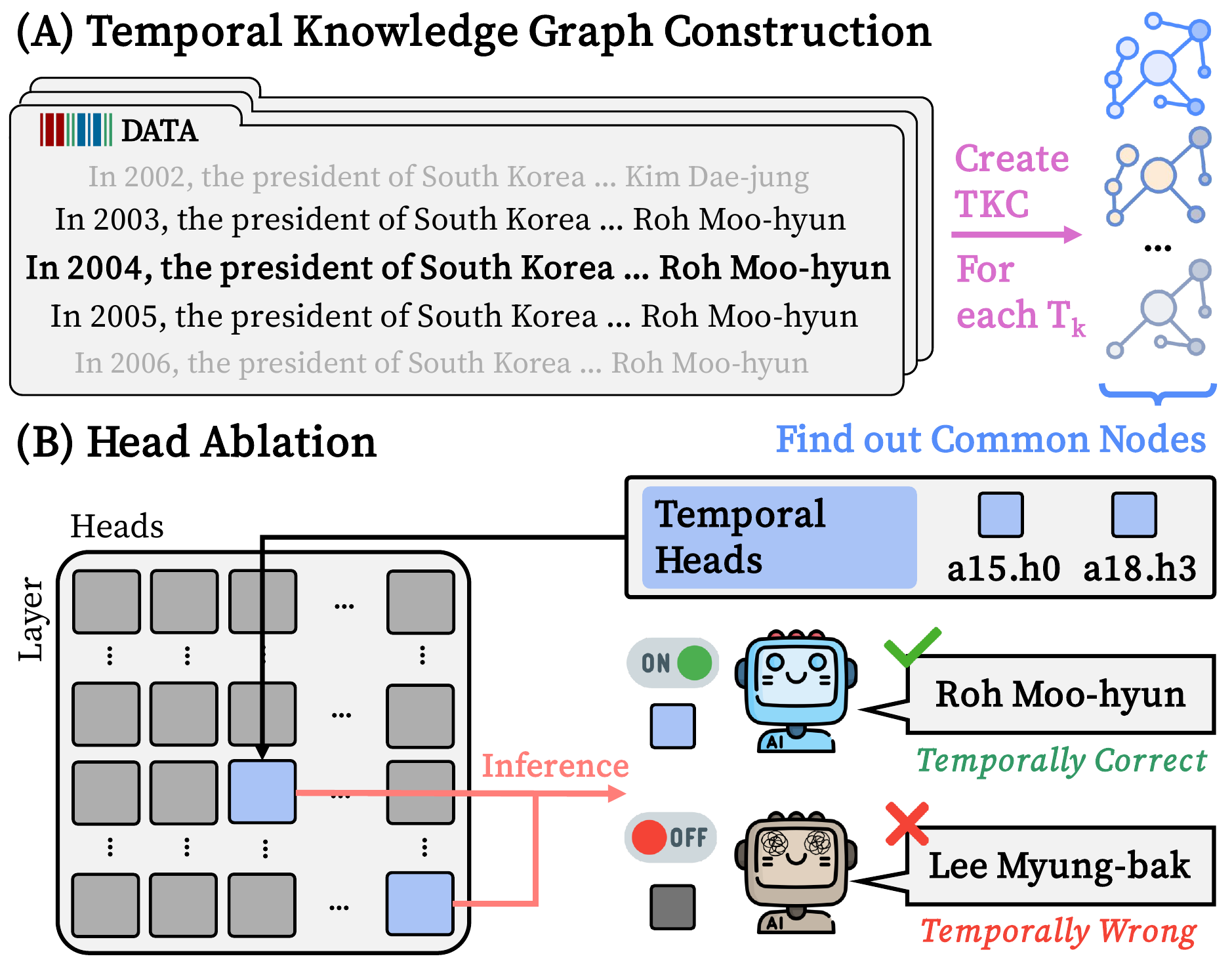}
\end{center}%
\vspace{-10pt}%
\caption{Temporal Heads exist within various TKCs at different times $T_k$.
Ablating them disrupts the model's temporal alignment, yielding incorrect objects.
}
\label{fig:intro_sample}
\vspace{-20pt}
\end{figure}

\begin{figure*}[t]
\vspace{-10pt}
\begin{center}
    \includegraphics[width=1\textwidth]{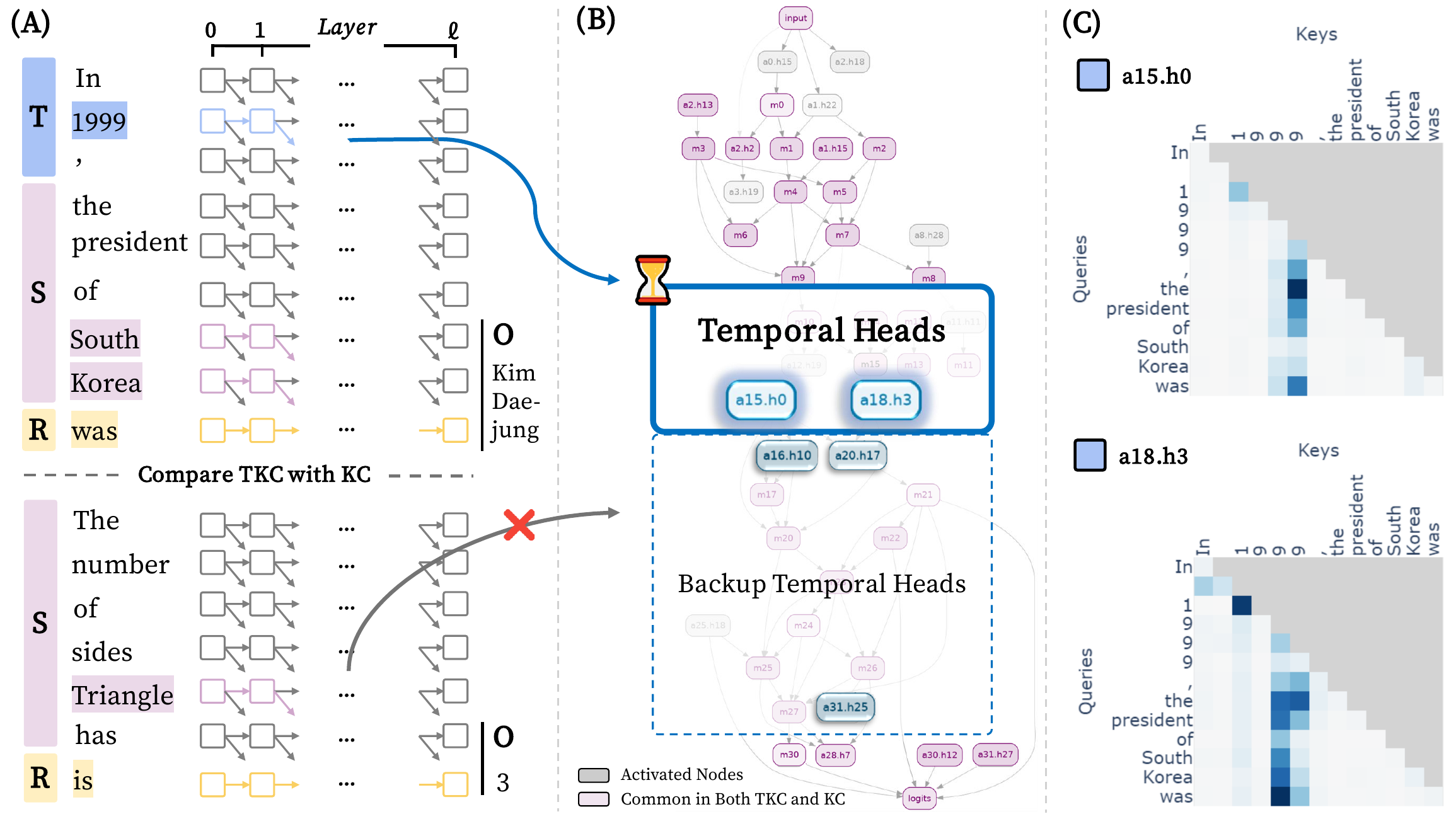}
\end{center}%
\vspace{-10pt}%
\caption{Overview of temporal knowledge circuit analysis.
(A): Construct temporal knowledge circuits (TKCs), and compare it with general knowledge circuits (KCs) using time-invariant knowledge.
Circuits reproduce residual streams for time~\textbf{\textcolor{cyan!70!gray}{T}}, subject~\textbf{\textcolor{pink!80!gray}{S}} and relation~\textbf{\textcolor{yellow!80!gray}{R}}.
This verifies temporal heads only found in each different TKCs of various year $T_k$.
(B): Example of simplified TKC.
Here, basic knowledge nodes is colored \textcolor{violet}{violet}, (common in both), while \textbf{\textcolor{blue!60!cyan!80!black}{Temporal Heads}} is highlighted.
(C): Attention map for Temporal Heads.
\textbf{a15.h0} means the 15th layer's first attention head.
Each head’s attention pattern is visualized with the attention weight assigned by the queries (row) to the keys (column).
Queries are the tokens distributing attention, and Keys are the tokens receiving attention.
Values represent attention weights, indicating the strength of this focus.
Total results are in Figures~\ref{fig:total_circuit}--\ref{fig:full_attn_phi}.
}
\label{fig:overview}
\vspace{-10pt}
\end{figure*}

Empirical observations suggest that LLMs already possess some level of temporal awareness~\citep{nylund2023time, dyknow}.
This raises the question of whether the model is inherently capable of encoding and utilizing temporal knowledge.
For instance, when prompted with time-specific queries like ``\textit{In 1999, [X] was a member of sports team}'', the model may generate the correct team \textit{[Y]} relevant to that year, indicating that certain time-conditional links are embedded in its internal parameters. 
The key puzzle, however, is how this temporal knowledge is organized and recalled.
Do LLMs internally have a place for \textbf{Time}, adjusting their factual outputs based on the input time condition?
If so, where within the model architecture—among the attention heads and feed-forward layers—does this mechanism reside?

To address them, we apply Circuit Analysis~\citep{ elhage2021mathematical, ioi} to reconstruct the model’s computations via localized subgraphs of attention heads, feed-forward networks, and residual streams.
Especially, by systematic ablating (zeroing out) attention heads or multilayer perceptron (MLP) components, it pinpoints which parts are responsible for eliciting knowledge in each recalling tasks~\citep{KC}.
These knowledge circuits enable to measure how much each nodes or edges in subgraph contribute to processing facts.

We extend it into temporal dimension, capturing how models reacts to time-evolving attributes using Temporal Knowledge Circuits (Figure~\ref{fig:intro_sample} (A)).
We then identify \textbf{\textit{Temporal Heads}}, such as \textit{a15.h0} and \textit{a18.h3}, which are exclusively activated for temporal knowledge while remaining inactive for time-invariant information.
Each model have its own temporal heads, which exhibit a strong influence on temporal input tokens in attention maps.
Moreover, ablating these heads significantly reduces time-specific factual accuracy, leading to temporal mismatches as suggested in Figure~\ref{fig:intro_sample} (B).

One step further, we explore in-depth impacts of temporal heads among different years, knowledge and conditioning types.
Ablating them exclusively affects temporal information, while having negligible impact on time-invariant knowledge and general question answering performance.
Notably, these temporal heads are activated for both numerical expressions (“In 2004”) and textual conditions (“In the year the Summer Olympics were held in Athens”), indicating that they encode a broader temporal dimension beyond simple numerical representation.
Building on this, we present that \textit{temporal knowledge editing}-selectively adding their activations-enables direct intervention in year-conditioned factual recall.
Through this targeted manipulation, our experiments demonstrate that the temporal heads serve as key subcomponents for encoding and modifying time-sensitive knowledge.

\section{Preliminaries}
\label{sec:preliminaries}
In this section, we provide background on the Circuit Analysis~\citep{zoom, circuitgrokking, conmy2023towards}, which represents the model's computation through structured subgraph of its components. 

\subsection{Circuit Analysis}
\label{subsec:circuit-analysis}
Circuit analysis represents a transformer's computation as a directed acyclic graph (DAG) $G = (N, E)$, where each node in $N$ corresponds to a distinct component in the model: attention heads $A_{l,j}$ (at layer $l$ and head $j$), MLP modules $M_l$ for each layer, the input node $I$ (embeddings), and the output node $O$ (logits). 
Thus, we formally define the set of nodes as:
\begin{equation}
    N = \{ I, A_{l,j}, M_l, O \}.
\end{equation}
The edges in $E$ represent residual connections that propagate activations between these nodes:
\begin{equation}
    E = \{ (n_x, n_y) \mid n_x, n_y \in N \}.
\end{equation}

A \emph{circuit} is defined as a subgraph $C \subseteq (N, E)$ selected to explain a specific behavior of interest--for instance, how certain tokens influence the model's output or how factual knowledge is stored and elicited.  
By examining which nodes and edges are crucial for producing a particular prediction, we can identify the subgraph (the circuit) that governs each behavior.

\subsection{Knowledge Circuit}
\label{subsec:knowledge_circuit}
A \emph{knowledge circuit}~\citep{KC} focuses on how a model treats the subject $s$, and relation $r$ to generate the object $o$ using a knowledge triplet $(s,r,o)$. 
By systematically \emph{ablating} (i.e.\ zeroing) parts of the model, it identifies the crucial nodes responsible for this generation and constructs a subgraph $KC \subseteq (N,E)$ whose removal \emph{breaks} the model’s ability to produce the correct object.
Concretely, it define a performance metric as:
\begin{equation}
\begin{split}
S(e_i) = &\; \log\bigl(p_G(o \mid s,r)\bigr) \\
         &- \log\bigl(p_{G/e_i}(o \mid s,r)\bigr).
\end{split}
\label{eq:knw-circuit-score}
\end{equation}
where $p_{G/e_i}$ denotes the model’s probability of next-token prediction after \emph{ablating} (i.e.\ zeroing) the activation of a node or edge $e_i$. 
If $S(e_i)$ exceeds a threshold $\tau$, $e_i$ is deemed \emph{critical} and retained in $KC$; otherwise, $e_i$ is pruned. 
This yields a \emph{minimal} set of heads/MLPs whose connections critically shape the binding of $(s,r)$ to the correct answer $o$.

Unlike a generic circuit for any functionality, a knowledge circuit specifically captures the local subgraph dedicated to storing and relaying factual content for the knowledge triplet at hand.
We specifically utilize effective attribution pruning-integrated gradients (EAP-IG), which ablating (zeroing) candidate edges and measuring drops in correct prediction~\citep{eapig}. 
For more details, see Appendix~\ref{sec:EAP-IG}.

\section{Knowledge Circuit Deciphers Temporal Head in LLMs}
\label{sec:knw-circuit-reuse}
We now explore how \emph{knowledge circuits}, extracted via EAP-IG pruning, can reveal specialized \emph{Temporal Heads} in large language models (LLMs). 
We extend knowledge circuits in \S \ref{subsec:knowledge_circuit} to 
\emph{temporal knowledge circuits} by analyzing how the same subject--relation pair can produce different objects across multiple time points. 
Specifically, we seek to identify which edges encode time-dependent specificity, such that an edge $e_i$ is crucial for predicting the time-relevant object $o_k$ at period $T_k$.
Given a knowledge circuit score $S(e_i)$ (Eq.~\ref{eq:knw-circuit-score}), 
we define its temporal variant as follows:
\begin{equation}
\begin{split}
S(e_i, T_k) = &\; \log p_G(o_k \mid s,r,T_k) \\
& - \log p_{G/e_i}(o_k \mid s,r,T_k) > \tau.
\end{split}
\end{equation}
where $T_k$ indicates a specific time (or period), and $o_k$ is the corresponding object for subject $s$ and relation $r$ at time $T_k$. 
Thus, $S(e_i, T_k)$ measures the contribution of edge $e_i$ to correctly predicting $o_k$ under time $T_k$.
For highlighting importance and simplifying graphs, edges retained in the temporal circuit satisfy $S(e_i, T_k) > \tau$, ensuring they encode time-dependent knowledge. 
Here, we decide to attach temporal conditioning in front of subject, following prior insight from causal tracing (\S\ref{sec:causal_tracing_theme}) and details in Appendix~\ref{sec:temporal-influence}.

\subsection{Implementations}
\label{subsec:impl}
We conduct experiments primarily on three LLMs: Llama-2-7b-chat-hf~\citep{llama2}, Qwen1.5-7B-Chat~\citep{qwen, qwen1.5}, Phi-3-mini-4k-instruct~\citep{phi3}.
We adopt transformer lens~\citep{transformerlens} to intercept and ablate model components, enabling \textbf{EAP-IG}-based circuit discovery.
We mainly illustrate results on Llama2, though similar trends emerge in the other models.
More details are described in Appendix~\ref{sec:detail_in_eapig}.

\subsubsection{Circuit Reproduction Score}
To evaluate how well a pruned circuit reproduces the full model’s behavior, we define the \emph{Circuit Reproduction Score} (CRS), ranging from $0$ to $100$. 
Let $B$ be the baseline performance of the full model on time-conditioned prompts, and $P$ be the performance of the pruned circuit. 
If the pruned circuit maintains or exceeds the baseline performance ($P \geq B$ when $B>0$), we assign it the maximum CRS as follows:
\begin{equation}
\mathrm{CRS}(B,P) = 100.
\end{equation}
Otherwise, the score follows an exponential decay:
\begin{equation}
\mathrm{CRS}(B,P) = 100 \times \sigma \exp\left(-\alpha \frac{d}{|B|} \right),
\end{equation}
where $d = \max\{B, 0\}$.
The factor $\sigma \in (0,1]$ accounts for sign mismatches, adjusting for cases where the pruned circuit’s output deviates in direction from the full model.
A higher CRS indicates better reproduction of the full model’s predictions. 
We describe the details of hyperparameters and adjustments to the Appendix~\ref{sec:detail_in_crs}.

\subsection{Dataset}
\label{subsec:dataset}
Our dataset comprises (statistics in Appendix~\ref{sec:dataset_details}):
\begin{itemize}
    \item \textbf{Temporal Knowledge}: Various categories of knowledge samples that embed a specific year (e.g., \emph{1999}, \emph{2004}, and \emph{2009}) alongside a factual statement (e.g., which sports team or president is correct in that year) based on Wikidata~\citep{wikidata}.
    \item \textbf{Time-Invariant Knowledge}: Commonsense data from LRE~\citep{lre} (e.g., \emph{object superclass}, \emph{fruit inside color}), plus newly implemented numerical facts embedded in subject/object (e.g., \emph{geometric shape} or \emph{roman numerals}). 
    These tasks assume no explicit time-based shift.
    \item \textbf{Unstructured QA}: We utilize TriviaQA~\citep{triviaqa} and Math~\citep{mathkg} QA in ChroKnowledge~\citep{chroknowledge} for unstructured, general QA to verify the ablation effect with basic LLM's tasks.
\end{itemize}
For each data point, we run both a \emph{clean} prompt and a \emph{corrupted} prompt, following EAP-IG guidelines.
We focus on the first token(s) that differ, capturing the key transition that determines correctness.
In the QA setting, we evaluate models using standard TriviaQA validation metrics, including exact match (EM) and F1 scores. 
For Math ChroKnowledge, we employ a multiple-choice QA (MCQA) template, scoring responses based on probability (\%). 
Given that models possess some degree of inherent knowledge~\citep{KC}, we assess their performance under zero-shot and greedy decoding.

\begin{table}[t]
\centering
\vspace{-5pt}
{\resizebox{\columnwidth}{!}{
\begin{tabular}{llllll}
\toprule
\multicolumn{2}{l}{\textbf{Category}} & \textbf{Knowledge} & \textbf{\#Node} & \textbf{\#Edge} & \textbf{CRS} \\ 
\midrule
\multicolumn{6}{l}{\textbf{\textit{Temporal}}} \\ 
\midrule
Sports     &            & Nicolas Anelka     & 29  & 37  & 74.14 \\ 
    &            & David Beckham       & 43  & 80  & 39.53 \\ 
Presidents &            & Argentina          & 42  & 102 & 60.97 \\ 
 &            & South Korea        & 46  & 110 & 65.55 \\ 
CEO        &            & Hewlett-Packard    & 52  & 115 & 53.49 \\ 
       &            & Chrysler           & 51  & 97  & 57.10 \\ 
Defense    &            & United States      & 50  & 137 & 48.08 \\ 
 &            & China              & 19  & 19  & 37.62 \\ 
\midrule
\multicolumn{3}{l}{\textbf{Avg}} & \textbf{42} & \textbf{87} & \textbf{54.56} \\ 
\midrule
\multicolumn{6}{l}{\textbf{\textit{Time-Invariant}}} \\ 
\midrule
CommonSense             &            & Object Superclass  & 43  & 56  & 44.47 \\ 
Conditional CS &            & Fruit Inside Color & 76  & 131 & 53.08 \\ 
Num in Obj     &            & Geometric Shape    & 52  & 118 & 76.09 \\ 
Num in Sub     &            & Roman Numerals     & 43  & 135 & 95.70 \\ 
\midrule
\multicolumn{3}{l}{\textbf{Avg}} & \textbf{54} & \textbf{110} & \textbf{67.33} \\ 
\bottomrule
\end{tabular}}}{}
\caption{Statistics of temporal knowledge circuits for Llama2, both temporal and time-invariant knowledge.
For temporal knowledge, each type of knowledge is reproduced with three selected years: \textbf{1999}, \textbf{2004}, and \textbf{2009}.
The numbers of nodes, edges and CRS is the average of each knowledge's yearly circuits.
}
\label{table:statistic_crs}
\vspace{-10pt}
\end{table}

\subsection{Evaluation}
\label{subsec:eval}
After pruning less-contributory nodes via EAP-IG, we measure how well the \emph{resulting subgraph} preserves the model’s original performance on each knowledge type. 
Table~\ref{table:statistic_crs} and~\ref{table:statistic_crs_qwen}--\ref{table:statistic_crs_phi} show the average number of \textbf{nodes} and \textbf{edges} in these pruned circuits, along with their \textbf{CRS}.
We then apply threshold \(\tau\) to remove edges/nodes that contribute marginally to object prediction, retaining only edges with scores above \(\tau\) and their corresponding nodes.

In Llama2, both temporal and time-invariant knowledge circuits effectively capture the model’s internal knowledge flow, with average CRS exceeding 50 in both cases. 
However, temporal circuits exhibit more variability, likely due to the inherent complexity of year-based facts. 
These tasks demand precise temporal conditioning, adding an extra difficulty, not just simply generating any possible objects. 
Even when models are expected to retain such knowledge, the increased complexity underscores the nuanced nature of temporal reasoning compared to time-invariant knowledge.

\subsection{Findings}
\label{subsec:findings}
We now identify \textbf{common nodes} in all circuits (e.g., \texttt{[input]}, \texttt{[logits]}, MLP \texttt{m2}, \texttt{m24}, \texttt{m30}, etc.) and a set of \textbf{temporal-only} nodes that appear exclusively in circuits for year-dependent prompts as in Figure~\ref{fig:overview}.
Firstly, most MLP nodes were appeared both temporal and time-invariant knowledge, as they are activated for storing knowledge~\citep{geva2021mlp, KN, whatKN}.

What stood out most was found in the attention heads.
\emph{\textbf{Temporal Heads}}, appearing in almost every temporal knowledge circuits but not time invariants, are shown: \verb|a15.h0|, \verb|a18.h3| in Llama2.
Those temporal heads reoccur across multiple year-specific circuits, and it is different for other model's cases like \verb|a17.h15| for Qwen 1.5 in Table~\ref{table:temporal_heads}.
Visualizing their attention maps in Figure~\ref{fig:overview} (C) indicates a strong focus on \emph{“In 19xx”} and subsequent subject phrases, as key tokens revolve around temporal conditions with queries hooking into the subjects.
This pattern corroborates the idea that these heads facilitate year-subject binding—justifying the label “temporal”, as this kind of task specific attention heads were previously suggested by~\citealp{ioi, circuitcolor, subhead, rethead} and ~\citealp{headsurvey}.

When lowering the ratio of exhibition (e.g., 70-80\%), additional heads (e.g., \verb|a0.h15|, \verb|a20.h17|, \verb|a31.h25|) emerge.
These \emph{Backup Temporal Heads} are also exclusive to temporal knowledge circuits, though their emerging varies different among types of knowledge and years.
But interestingly, even at high ratio, no heads are exclusive in time-invariant knowledge circuits.
This suggests that many “general knowledge” heads overlap with or are reused by knowledge recalling tasks, whereas certain specialized heads exist \emph{only} for time-based tasks.

\begin{figure*}[t]
\vspace{-10pt}
\begin{center}
    \includegraphics[width=1\textwidth]{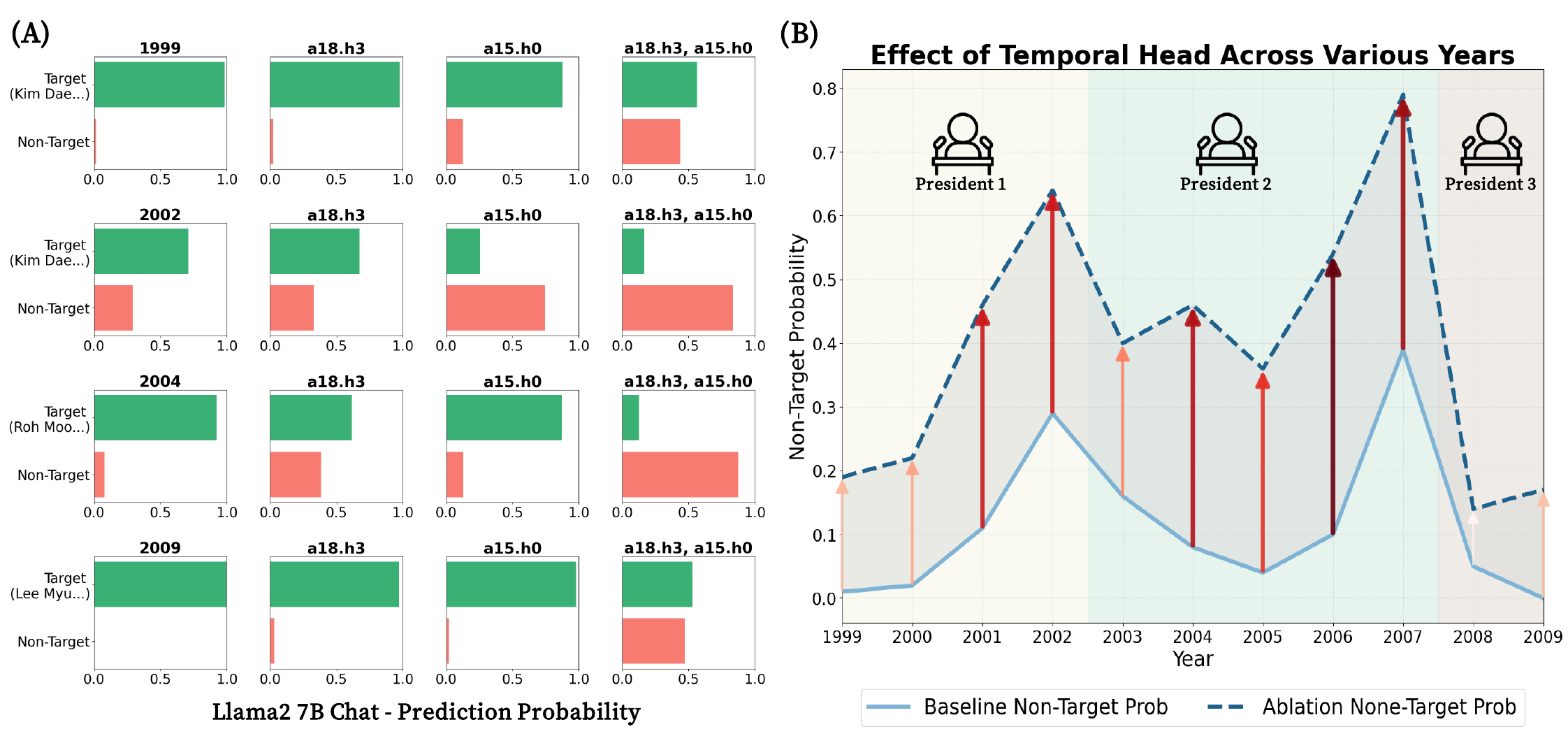}
\end{center}%
\vspace{-10pt}%
\caption{Log probability results with temporal knowledge; \textit{In XXXX, the president of South Korea was}.
(A) shows prediction probability change among results of Llama2.
The effect of head ablation reacts differently for each selected year with the same prompt.
Each subplot in (A) represents the probability distribution of correct (\textcolor{green!60!blue}{green}) and incorrect (\textcolor{red!80!white}{red}) predictions, where the x-axis denotes probability values and the y-axis differentiates between target and non-target responses.
Total results for each model are in Figures~\ref{fig:log_prop_app1}--\ref{fig:log_prop_app2} in Appendix.
(B) illustrates the performance degradation trends across various years.
As averaging the result of ablation, the \textcolor{gray}{\textbf{gray space}} between two line plots represent degradation level pointed out by \textcolor{red!70!black}{\textbf{red arrows}} (which becomes darker and bigger when the gap is wider).
The background shows how objects were changed in the time range between 1999 to 2009.}
\label{fig:log_prob}
\vspace{-10pt}
\end{figure*}

In the next (\S\ref{sec:indepth-analysis}), we delve into further ablation experiments to verify that ablating temporal heads indeed degrades year-specific predictions, reinforcing their role as the crucial channel through which the model recall knowledge conditioned on time.

\section{In-Depth Analysis of Temporal Heads}
\label{sec:indepth-analysis}
We conduct a more fine-grained analysis to understand \emph{how} temporal heads identified in the extracted circuits impact final predictions, especially for temporally changing facts. 
Drawing inspiration from~\citealt{logprob} on \emph{log-probability} based evaluation, we perform targeted \emph{Attention Head Ablation Inference} (\S\ref{subsec:head-ablation}) to observe how the model’s confidence shifts when certain “temporal” heads are zeroed out. 
We then test an \emph{Alias} scenario with temporal conditioning in textual context (\S\ref{subsec:alias-test}) to see if the same heads reappear for less explicit time references. 
Finally, we explore a \emph{Temporal Knowledge Editing} (\S\ref{subsubsec:temp-edit}) that uses attention addition to reinforce or awake year-specific content.

\begin{table}[t]
\centering
{\resizebox{\columnwidth}{!}{
\begin{tabular}{lllll}
\toprule
\textbf{THs} & \textbf{Settings} & \textbf{Temporal (\%)} & \textbf{Invariant (\%)} & \textbf{QA (F1)} \\
\midrule
\rowcolor[HTML]{BFD9EC}
\multicolumn{5}{l}{\textbf{\textit{Llama-2-7b-chat-hf}}} \\
\midrule
\multirow{2}{*}{\textbf{a18.h3}, \textbf{a15.h0}} & Baseline & 29.7 & 61.8 & 55.4 \\
                                  & Ablation & 25.6~\textcolor{red}{\(\Downarrow\)} & 61.7 & 54.9 \\
\midrule
\rowcolor[HTML]{D4BFE1}
\multicolumn{5}{l}{\textbf{\textit{Qwen1.5-7B-Chat}}} \\
\midrule
\multirow{2}{*}{\textbf{a17.h15}}        & Baseline & 22.4 & 62.7 & 49.7 \\
                                  & Ablation & 19.8~\textcolor{red}{\(\Downarrow\)} & 62.6 & 49.5 \\
\midrule
\rowcolor[HTML]{8EDB8A}
\multicolumn{5}{l}{\textbf{\textit{Phi-3-mini-4k-instruct}}} \\
\midrule
\multirow{2}{*}{\textbf{a10.h13}}        & Baseline & 35.4 & 59.8 & 46.8 \\
                                  & Ablation & 26.0~\textcolor{red}{\(\Downarrow\)} & 60.6 & 46.2 \\
\bottomrule
\end{tabular}}}{}
\caption{Temporal Heads (THs) across different LLMs. 
The scores besides each heads are evaluated in three cases (temporal knowledge, time-invariant knowledge, and TriviaQA) with two settings (baseline inference and ablation inference).
Scores are checked with the average performance for each tasks, measured in probability (\%) or F1 score. 
While performance in temporal knowledge drops significantly (3 to 9\%), time-invariant and general QA remain relatively stable or even goes up.}
\label{table:temporal_heads}
\vspace{-10pt}
\end{table}

\subsection{Attention Head Ablation Inference}
\label{subsec:head-ablation}
\paragraph{Motivation}
While temporal knowledge circuit construction based on EAP-IG pruning (\S\ref{sec:knw-circuit-reuse}) reveals the structure of temporal knowledge processing, we still need direct evidence that certain “temporal heads” genuinely mediate year-based predictions. 
We adopt a \emph{hard-coded} approach that sets the selected attention head’s output weights to zero, thus preventing it from contributing to the residual stream. 
We then measure changes in the model’s log probability for the correct target object vs.\ competing objects in different time.
\paragraph{Log Probability Variation}
Following \citealt{logprob}, we assess temporal knowledge retention by evaluating changes in object probabilities under head ablation.  
Let \({O}\) be the set of all candidate objects (e.g., teams, presidents) in the time range, and \( p(o | s, r, T) \) the model’s probability of selecting object \( o \) from subject $s$, relation $r$ and time $T$.
The model’s default choice is labeled \texttt{Target} if it matches the correct temporal fact, otherwise \texttt{Non-Target}.  
After ablating suspected temporal head(s), we recompute object probabilities:

\begin{align}
    z_o &= \log p_\text{ablate}(o|s,r,T),
    \\ \hat{p}_o &= \frac{\exp(z_o)}{\sum_{o' \in O} \exp(z_{o'})},
\end{align}
where \( p_{\text{ablate}} \) denotes the log-probability computed by forward pass of model, ablating corresponding heads. 
This evaluates how the probability distribution over \( O \) shifts, rather than just predicting the most likely answer.
Details in Appendix~\ref{sec:log_evaluation_details}.

\subsubsection{Result of Temporal Knowledge}
As shown in Figure~\ref{fig:log_prob} (A), ablation significantly reduces log probability for the correct year-specific \texttt{Target} in temporal tasks.
When ablating \verb|a15.h0| or \verb|a18.h3| or both of them, the model frequently chooses \texttt{Non-Target} objects from \({O}\) (e.g., a president of different year). 
Not just raising of those percentage, specific attention heads influence each years differently; some are more critical for 1999, while others have a stronger effect in 2004 or 2009.
For instance, ablating \verb|a18.h3| significantly impacts 2004 but has a lesser effect on 2002.  

Figure~\ref{fig:log_prob}~(B) illustrates the varying degrees of performance degradation across different years. 
The \textbf{red arrows} highlight these degradation levels, where darker and thicker arrows indicate a more pronounced effect of ablation. 
Notably, around \textbf{object transition periods} (e.g., between 2002--2003 and 2007--2008), the non-target probability spikes, confusing when knowledge boundaries shift along the timeline. 
This aligns with the intuition that temporal knowledge transitions introduce uncertainty in the model’s predictions in temporal context.

\subsubsection{Result of Time Invariant Knowledge}
\begin{figure}[t]
\vspace{-10pt}
\begin{center}
    \includegraphics[width=\columnwidth]{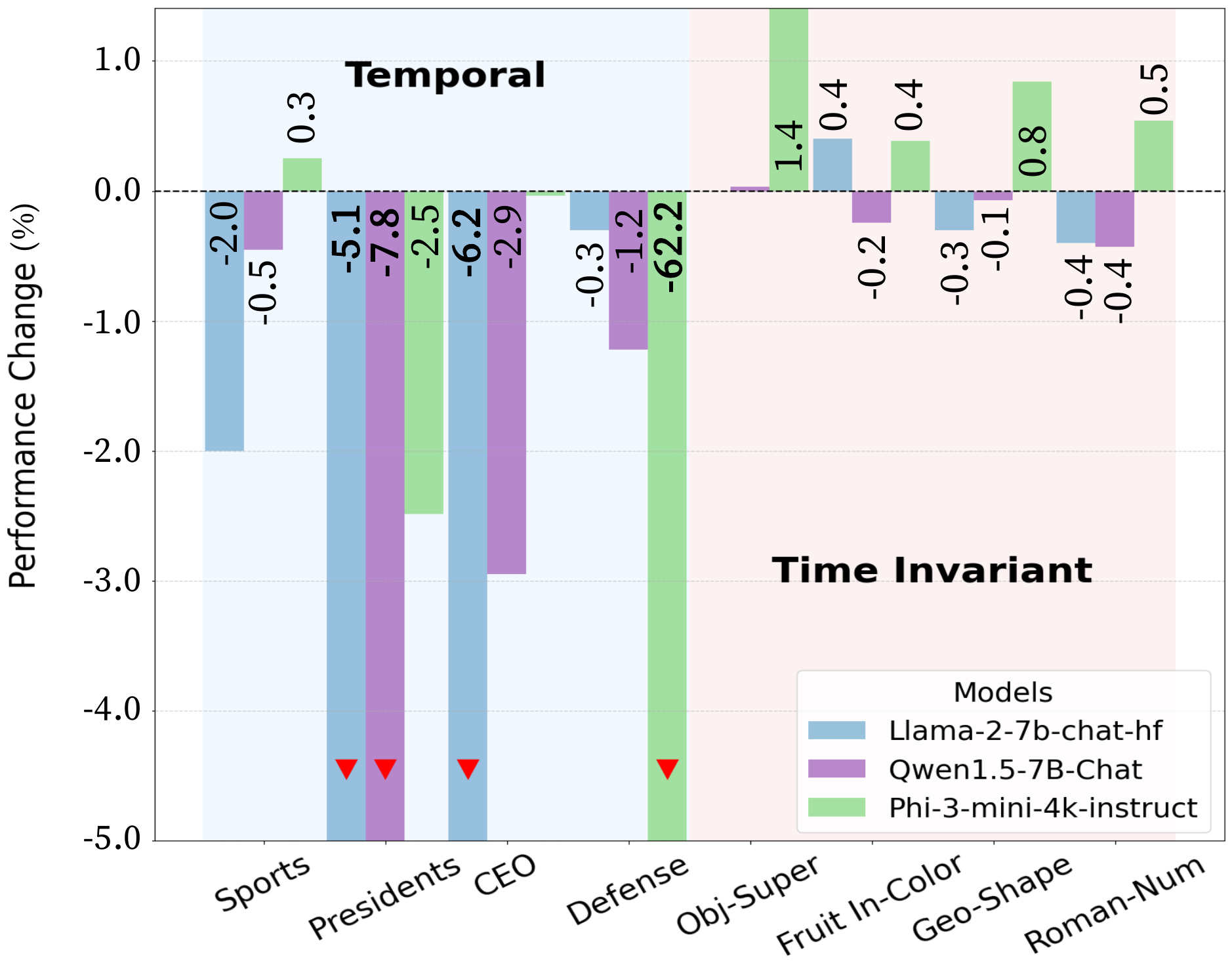}
\end{center}%
\vspace{-10pt}%
\caption{
Head ablation effect across various knowledge types.
Three selcted model shows distinct differentiation for temporal knowledge (\textcolor{cyan!70!gray}{left side}) and time invariant knowledge (\textcolor{pink!80!gray}{right side}).
The change of performance is calculated with the average score of baseline (non-ablation) and modified (ablated result), using model specific temporal head information.
While degrees of degradation is different among models, overall tendency reflects the importance of temporal head to inference temporal knowledge.
}
\label{fig:ablation}
\vspace{-10pt}
\end{figure}

By contrast, ablating the same heads for \emph{invariant} knowledge (e.g., \emph{fruit inside color}) causes minimal performance drop in Table~\ref{table:temporal_heads} and Figure~\ref{fig:ablation}.
This indicates that “temporal heads” indeed route only temporally conditioned knowledge, and disabling them forces the model to make temporally incorrect rather than incorrect of stable knowledge.
Besides, Phi-3-mini-4k-instruct affects more sensitively than others as its parameter size is half of other two models, resulting more reactive to small changes in attention alignment.
This even causes a slight gain of performance in time-invariant knowledge tasks.

\subsubsection{Result of General QA}
As Table~\ref{table:temporal_heads} and result in Appendix~\ref{app:total_qa} shown that ablating temporal heads doesn't harm common knowledge recalling or answering general knowledge questions.
Here, we test TriviaQA and Math ChroKnowledge and find out that just ablating temporal heads doesn't affect the performance of basic QA, droping almost less than 0.6 in F1 score.

\subsection{Alias Test With Textual Conditioning}
\label{subsec:alias-test}
In previous findings of Section~\S\ref{subsec:findings}, we experimented with cases where numeric values were present either in the prompt (\textit{Roman Numerals}) or in the answer object (\textit{Geometric Shape}) under time-invariant conditions (like “\emph{Triangle has 3 sides}”).
For all scenarios, temporal heads did not emerge, suggesting that their activation is not merely a response to numerical information but rather specific to temporal knowledge processing.
We further investigate whether these same heads appear for less direct numeric conditioning.
Instead of a literal \textit{“In 2004”} prompt, we use “\textit{In the year the Summer Olympics were held in Athens}” or “\textit{For his first},” providing an \emph{indirect} textual condition referencing the relevant time. 
We again construct knowledge circuits and observe which heads surpass threshold.

Such “alias” statements yield smaller CRS (e.g., 40.3 in president cases), though, temporal heads still appears.
These heads may not always exceed normal threshold (e.g.\ \(\tau = 0.1\)), they still register moderate importance. 
Coupled with results from the numeric “In 2004” prompt, this indicates that those heads do \emph{not} rely solely on numeric tokens, but also respond—albeit less strongly—to textual or event-based temporal conditioning.
This further validates that they encode a \emph{temporal} dimension, rather than merely responding to arbitrary numbers.
Visualized results are in Figure~\ref{fig:alias_app} of Appendix.

\begin{table}[]
\vspace{-10pt}
\centering
{\resizebox{\columnwidth}{!}{%
\begin{tabular}{lccccccccc}
\toprule
\multicolumn{1}{c}{\multirow{2}{*}{\textbf{Set}}} & \multicolumn{7}{c}{\textbf{Temporal Knowledge (\%)}} & \multicolumn{1}{c}{\multirow{2}{*}{\textbf{Avg}}} \\ \cmidrule(lr){2-8}
\multicolumn{1}{c}{} & \multicolumn{1}{c}{\textbf{Spo}} & \multicolumn{1}{c}{\textbf{Prez}} & \multicolumn{1}{c}{\textbf{CEO}} & \multicolumn{1}{c}{\textbf{Def}} & \multicolumn{1}{c}{\textbf{Mov}} & \multicolumn{1}{c}{\textbf{GDP}} & \multicolumn{1}{c}{\textbf{Infla}} & \multicolumn{1}{c}{} \\ \midrule
\multicolumn{9}{l}{\textbf{Fundamental Prompt}: \textit{In XXXX, Lionel Messi was a member of ...}} \\ \midrule
Base & 41.9 & 80.7 & 27.5 & 13.5 & 23.1 & 10.4 & 10.8 & 29.7 \\
Abl & \textcolor{red}{40.0} & 75.6 & 21.3 & 13.3 & \textcolor{red}{9.37} & 10.7 & 9.34 & 25.6 \\ \midrule
\multicolumn{9}{l}{\textbf{Prompt Variation}: \textit{In year XXXX, Lionel Messi was a member of ...}} \\ \midrule
Base & 40.5 & 82.6 & 45.8 & 13.6 & 22.1 & 17.3 & 12.1 & 33.4 \\
Abl & 39.7 & \textcolor{red}{75.4} & 43.2 & \textcolor{red}{13.2} & 14.5 & \textcolor{red}{14.1} & 10.3 & 30.0 \\ \midrule
\multicolumn{9}{l}{\textbf{Real-World Question}: \textit{In XXXX. which sports team was Lionel ...}} \\ \midrule
Base & 40.7 & 81.5 & 55.6 & 10.1 & 24.1 & 19.2 & 10.5 & 34.5 \\
Abl & 40.5 & 74.8 & \textcolor{red}{42.6} & 10.1 & 16.4 & 17.8 & \textcolor{red}{8.76} & \textcolor{red}{30.1} \\ \bottomrule
\end{tabular}}}{}
\caption{Results of prompt variations, comparing baseline performance to the ablated model (i.e., removing Temporal Heads). 
Categories of temporal knowledge are same as Table~\ref{table:total_result1}, which is Sports, Presidents, CEO, Defense, Movies, GDP and Inflations.
Each scores were measured in probability (\%) with averaging effect of multiple heads ablation results (a15.h0 and a18.h3 for Llama2).
The most dropped score for each column is colored red.}
\label{table:rebuttal}
\vspace{-10pt}
\end{table}

\subsection{Additional Test With Prompt Variations}
\label{subsec:prompt-variation}

For more qualitative experiments, particularly regarding prompt settings, we do more analysis study with various prompt styles.
As we originally used only one fundamental prompt format (e.g., \textit{"In XXXX"}), we have conducted additional ablation experiments with Llama2 under new prompt settings to address concerns about generalizabilty.

\begin{itemize}
    \item \textbf{Variation of Fundamental Prompt} maintains the core temporal format while adding slight textual variation (\textit{“In year XXXX”}).
    \item \textbf{Real-World Question Format} simulates a more practical Q\&A scenario, moving beyond a declarative statement into a direct question.
\end{itemize}

We evaluate each prompt settings on multiple knowledge categories in Table~\ref{table:rebuttal} same as Table~\ref{table:total_result1} in Appendix.
Here, we find out key observations.
Across all cases, ablating Temporal Heads (i.e., disabling them) yields a notable drop in performance for each knowledge category.
So, even when the prompt style changes from the original \textit{“In XXXX”} to \textit{“In year XXXX”} or a question-based format, the importance of Temporal Heads remains evident.
These results reinforce our claim that Temporal Heads underlie the model’s ability to handle time-sensitive knowledge, regardless of prompt diversity, demonstrating our approach's generalizability.

\subsection{Difference with General Formats}
\label{subsec:difference-generalkcs}

Interestingly, if we suggest prompt without explicit temporal aspects such as \textit{"Who is the president of South Korea?"} and \textit{"The president of Argentina is"}, the results indicated no meaningful nodes or edges in every circuits.
Using Llama2, we conduct two approaches to addressing these scenarios:

\begin{itemize}
    \item Construct circuits by comparing target object with objects from different years as corrupted run (Isolating only the temporal component).
    \item Construct circuits by comparing target object with objects from different knowledge contexts as corrupted run (e.g., presidents of different countries).
\end{itemize}

In each results, CRS dropped significantly (e.g., 61 to 18 in same knowledge category), reflecting weak or absent circuit reconstruction in scenarios lacking explicit temporal conditions.
Still, ablating Temporal Heads affected the model's greedy decoding outputs (e.g., switching the answer from "Moon Jae-in" to a temporally alternative object in different year "Park Geun-hye" in response to prompt “The president of South Korea is”). 
This implies that such queries, despite appearing static, inherently involve subtle temporal components.
Thus, we conclude that it is challenging to construct temporal knowledge circuit without temporal conditions as it is inherently temporal, and it is hard to distinguish model’s activation difference between objects without temporal aspects, which needs to construct circuits. 
This finding reinforces the validity of our methodology—isolating the unique characteristics of temporal knowledge through comparison with commonsense knowledge.
\section{Temporal Knowledge Editing}
\label{subsubsec:temp-edit}
\begin{figure}[t]
\vspace{-10pt}
\begin{center}
    \includegraphics[width=\columnwidth]{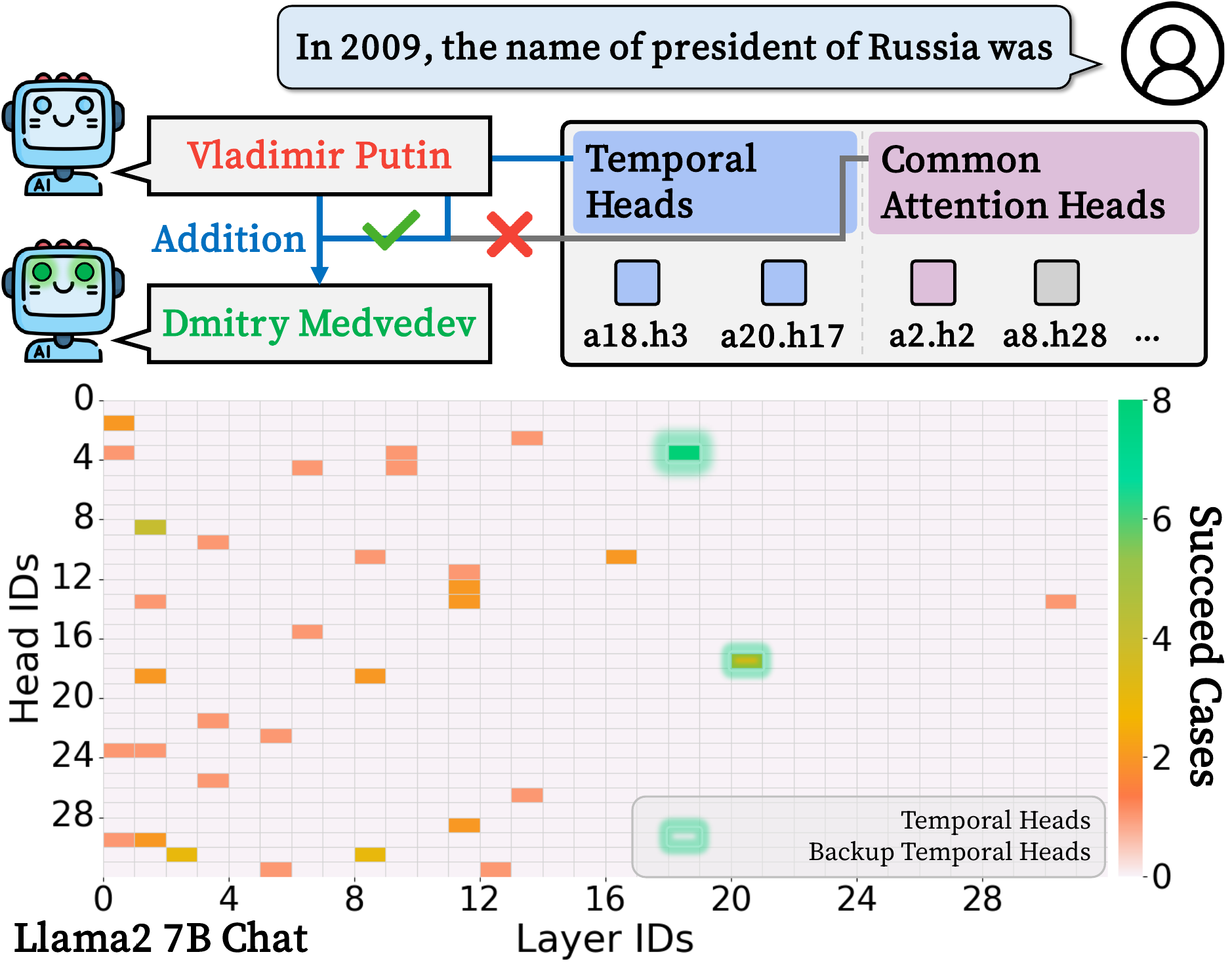}
\end{center}%
\vspace{-10pt}%
\caption{Example of Temporal Knowledge Editing.
From the source prompt, we catch the specific attention value of model's head, for example, \textbf{a18.h3}.
By simply adding it to target prompt, the model's output is changed into temporally correct answer from temporally wrong answer.
The headmap below denotes the number of success in editing for every combination of layers and heads.
The most successful case in here is temporal heads \textbf{a18.h3} as highlighted, following other heads such as backup temporal heads \textbf{a20.h17}.
}
\label{fig:editing}
\vspace{-10pt}
\end{figure}

Lastly, we explore an approach to confirm that injecting or amplifying \emph{temporal head}'s attention value can effectively \emph{“edit”} year-specific knowledge as in Figure~\ref{fig:editing}. 
Given a \texttt{source\_prompt} (where the model is confident about a certain year’s fact) and a \texttt{target\_prompt} (where it confuses the same year) based on log probability results, we:
\begin{enumerate}
    \item \textbf{Extract} the value of attention head \(\mathbf{a}_{\mathrm{src}}\) from the \texttt{source\_prompt} at a chosen layer/head (e.g.\ \verb|a18.h3|).
    \item \textbf{Average} over total source prompts (e.g., "In 2009, the name of president of South Korea was").
    \item \textbf{Inject} the modified attention value into the \texttt{target\_prompt} at the corresponding temporal token position, scaled by a coefficient \(\lambda\):
\end{enumerate}
Details of adding an attention is in Appendix~\ref{appendix:temp-edit}.

This modification is applied dynamically using a forward hook mechanism at inference time, preserving the overall model parameters while selectively influencing time-conditioned factual recall.  
We test it with model wrong answer in a normal condition, varying the injection coefficient across three cases (\(\lambda = 1, 3, 6\)), following~\citealp{actadd, actaddllama2}, which emphasized its impact.

Remarkably, the model’s completions shift from a temporally incorrect response (\textit{“changed to Vladimir Putin”}) to the correct one (\textit{“Dmitry Medvedev”}), aligning with the known presidency timeline.
The heatmap in Figure~\ref{fig:editing} further supports this by visually representing the effectiveness of temporal knowledge editing across all layers and heads.  
While certain attention heads can influence the model’s response, the \textbf{most successful cases} are consistently linked to temporal heads, with \verb|a18.h3| exhibiting the highest success rate.  
Additionally, backup temporal heads, such as \verb|a20.h17|, also rank among the top-performing heads, reinforcing their critical role in preserving and modifying time-conditioned knowledge.  
This highlights that temporal factual recall is not arbitrarily distributed but is instead concentrated in specialized subcomponents.
Other results are in Figure~\ref{fig:editing_app}.

This targeted intervention remains minimally invasive, as it does not require global fine-tuning but instead modulates the value of a single specialized head, thereby preserving most of the model’s prior knowledge.  
Taken together, these findings reinforce the hypothesis that LLMs harbor a \emph{temporal subcomponent} within specialized attention heads.  
By intercepting or amplifying these temporal heads, we can selectively alter time-conditioned responses, strengthening the claim that these heads are integral to the reinforcement of year-based factual knowledge.

Furthermore, we applied attention addition-based temporal knowledge editing to those prompt variations in Section~\ref{subsec:prompt-variation}. 
Although the number of successful edited cases decreased approximately by half in the Llama2 model compared to the fundamental setting (from 8 to 4), the head with the highest number of success remained the Temporal Head a18.h3. 
This demonstrates that attention addition is also effective across diverse prompt settings, highlighting potential synergies through integration with other methodologies to further enhance temporal knowledge processing capabilities.
\section{Related Works}
\label{sec:related-works}

\subsection{Temporal Knowledge of LLM}  
Despite advancements in LLMs, handling \emph{temporal knowledge} remains a key challenge. 
While prior works focus on factual consistency~\citep{lama, negated} or refining model editing in MLP layer~\citep{mend, rome, memit}, few address how facts evolve over time. 
Studies on time-aware QA and temporal probing~\citep{timeqa, zhang2021situatedqa, dhingra-etal-2022-time, temporalwiki} reveal that LLMs struggle with dynamically shifting facts.
Recent approaches attempt explicit temporal alignment~\citep{carpediem, settheclock, dyknow, chroknowledge}, but have focused on external evaluations.

Beyond these approaches, interpretable studies have mathematically and empirically demonstrated that LLMs can implicitly interpolate and process continuous temporal and spatial information, as well as compositional heuristics, even without explicit context in the training corpus~\citep{implicitly, Arithmetic}.
Building on this, our findings highlight that LLMs encode temporal facts implicitly, relying on manipulable attention heads, underscoring the need for better temporal supervision and disentangled knowledge representations.

\subsection{Attention Heads in Language Models}
\label{subsubsec:attention-heads}
Under mechanistic interpretability~\cite{zoom, causal, open}, researches about attention heads were done by~\citealp{syntactichead, ioi, copysuppression}, showing off specific heads that copy key tokens to the output, ensuring consistency in transformers.
These \textit{Mover Heads} are a kind of induction heads~\citep{inductionhead} moving syntactic information~\citep{subwordmergehead}.
Other works were followed as finding out retreval heads~\citep{rethead}, heads for semantic information for color~\citep{circuitcolor}, or subject and relation~\citep{subhead}.
Those various kinds of attention heads attend to critical tokens and directly influence the logits by writing their embeddings into the residual stream~\citep{headsurvey}.

Experiments show that ablating those heads significantly disrupts tasks like syntactic induction or semantic information understanding, highlighting their specific roles.
A special case, \emph{Backup Heads}, remains inactive under normal conditions but replicates task specific head functionality when primary heads are ablated. 
This ensures model robustness by maintaining token copying behavior even when key circuit components are disrupted.
We treat founded temporal attention heads as a subcategory of semantic heads like subject heads and relation heads~\citep{subhead} in our experiments.

\section{Conclusion}
\label{sec:conclusion}
We systematically investigate how LLMs can handle \emph{temporal knowledge}, focusing on time-dependent facts. 
Through our experiments, we uncovered \emph{Temporal Heads} that selectively mediate the activation of time-variant knowledge.
Ablating these heads leads to temporal mismatches while leaving time-invariant knowledge and general QA performance unaffected.
Note that these heads are also activated under textual conditioning, and using their value for editing successfully changes the models' responses with minimal intervention.

As a foundational step, our work explores how LLMs can actively manage temporal information rather than merely integrating temporal context.
We believe our analysis offers valuable insights into the inner mechanisms of LLMs and can inspire future approaches for \emph{time-aware model alignment} and \emph{precise temporal updates} by selectively targeting \emph{temporal heads}, rather than relying on global retraining. 

\section*{Limitations}
While our approach demonstrates promising results in identifying and analyzing temporal knowledge circuits, we acknowledge some limitations in our current work. 

First, analysis of unstructured temporal QAs like General ChroKnowledge~\citep{chroknowledge} were constrained, as the underlying multiple-choice options in those tasks typically do not exhibit temporal dependencies.
So we focused more on our temporal knowledge dataset, abundantly describing the effect of ablation in these cases.
In addition, even though our approach systematically validated the importance of Temporal Heads in processing temporal knowledge, future work should include its broader application for enhancing temporal reasoning to increase practical value.

On the other side, as EAP-IG didn't support models with Grouped-Query Attention (GQA), which cannot use the \texttt{split\_qkv\_input} option, our main analysis exclude those models like Llama-3-8B-Instruct~\citep{llama3}.
Still, we checked their results and found that even their CRS is not quite enough and their circuit construction is not detailed, temporal heads are still could be founded: \emph{a18.h15} and \emph{a23.h26}.

\subsection*{Acknowledgments}
This work was supported in part by the National Research Foundation of Korea [NRF-2023R1A2C3004176, RS-2023-00262002], the Ministry of SMEs and Startups [RS-2024-00523644], the Ministry of Health \& Welfare, Republic of Korea [HR20C002103], and the ICT Creative Consilience program through the Institute of Information \& Communications Technology Planning \& Evaluation (IITP) grant funded by the MSIT [IITP-2025-RS-2020-II201819].

\bibliography{custom}

\clearpage
\section*{Appendix}
\
\appendix
\section{Effective Attribution Pruning-Integrated Gradients}
\label{sec:EAP-IG}
We perform \textbf{Effective Attribution Pruning} (EAP) by ablating (zeroing) candidate edges and measuring the drop in correct predictions following~\citealp{eapig}. 
In tandem, we use \textbf{Integrated Gradients} (IG) to capture gradient-based contributions:
\begin{equation}
\mathrm{IG}(\mathbf{z},\mathbf{z}') = \int_{0}^{1} 
\frac{\partial}{\partial \mathbf{z}} \mathcal{L}(\mathbf{z}' + \alpha(\mathbf{z}-\mathbf{z}'))
\,\mathrm{d}\alpha,
\end{equation}
where $\mathcal{L}$ is the loss (e.g., negative log-likelihood), and $\mathbf{z}'$ a baseline embedding or activation. 
Furthermore, not just combining signals to rank each node/edge by its importance, we extend EAP-IG to \emph{time-sensitive} knowledge.
We construct \textbf{temporal knowledge circuits} by analyzing variations across different years $T_k$. 
For a given $(s,r)$ pair:
\begin{itemize}
    \item \textbf{Clean input}: $(s,r,o_t)$ where $o_t$ is correct at $T_t$.
    \item \textbf{Corrupted inputs}: $(s,r,o_{t'})$ where $o_{t'}$ is the correct object for a different time $T_{t'} \neq T_t$.
\end{itemize}
Rather than treating $o_{t'}$ as incorrect, we leverage the contrast between different valid temporal associations to isolate time-dependent components. An edge $e_i$ is retained in the temporal circuit if:
\begin{equation}
\begin{split}
S(e_i, T_k) = &\; \log p_G(o_k \mid s,r,T_k) \\
& - \log p_{G/e_i}(o_k \mid s,r,T_k) > \tau.
\end{split}
\end{equation}
This identifies edges that encode temporal specificity rather than general factual associations.
By ablating edges across different $T_k$, we verify if disruptions occur primarily at the corresponding time while preserving outputs for other years. This ensures the extracted circuits genuinely reflect temporal dependencies.

\subsection{Implementation Details in EAP-IG}
\label{sec:detail_in_eapig}
In each model’s configuration, we set \texttt{split\_qkv\_input} to true in transformer lens~\citep{transformerlens}, ensuring attention heads are disentangled enough for targeted pruning.
The \texttt{ig\_steps} for integrated gradients, we set it as 100.
We use \texttt{top\_n} 5000 settings and the $\tau$ for simplified threshold, we use 0.1 as a predefined value for every models to cutting out unimportant edges and nodes.
The experiments are all done with one NVIDIA A100 GPUs (80GB), less than 30 minutes per each runs.

\section{Causal Tracing}
\label{sec:causal_tracing_theme}
Causal Tracing~\citep{causal, rome} aims to reveal which hidden states in an autoregressive Transformer \emph{cause} correct recall of a fact. 
Let a fact be $(s,r,o)$ (e.g., $(\texttt{L. Messi}, \texttt{sports\_team}, \texttt{Newell's Old Boys})$), and time $T$ (e.g., $\texttt{In 1999}$). 
We construct a prompt $p$ (e.g., \textit{“In 1999, Lionel Messi was a member of sports team …”}) and measure the model’s probability of generating $o$ at output:
\begin{equation}
p_{\mathrm{clean}}(o) = G(p),
\end{equation}
where $G$ is the Transformer. Next, we create a \emph{corrupted} prompt $p^\prime$ (e.g., replacing “Lionel Messi” with a fake name). Denote the model’s probability,
\begin{equation}
p_{\mathrm{corr}}(o) = G(p^\prime).
\end{equation}
Because key information is obfuscated, $p_{\mathrm{corr}}(o)$ typically drops. 
Finally, in the \emph{corrupted-with-restoration} run, we overwrite certain hidden states in the corrupted run with their clean-run counterparts:
\begin{equation}
p_{\mathrm{restored}}(o) = G_{\mathrm{restore}}\Bigl(p^\prime,\{\mathbf{h}^{(l)}_{\mathrm{clean}}\}\Bigr),
\end{equation}
where $\mathbf{h}^{(l)}_{\mathrm{clean}}$ are layer-$l$ hidden states from the clean run. 
If restoring layer $l$ significantly boosts $p_{\mathrm{restored}}(o)$, those states at layer $l$ are \emph{causally important} for retrieving the fact. 
Applying this procedure to time-conditioned facts (e.g., specifying “In 1999,” “In 2009,” etc.) localizes \emph{temporal} knowledge within specific tokens and layers.
\section{Where Does Temporal Condition Exert Influence on Knowledge Triplets?}
\label{sec:temporal-influence}

We next investigate precisely \emph{where} a temporal cue, such as \textit{“In 1999,”} or \textit{“In 2004,”} exerts its main influence within the triplet $(s,r,o)$. 
To this end, we adopt a causal-tracing approach (inspired by ROME~\citep{rome}) targeted at isolating \emph{temporal} effects. 
Specifically, we compare two prompts:

\begin{itemize}
    \item \textbf{Without Temporal Cue:} \emph{“Lionel Messi was a member of sports team ...”}
    \item \textbf{With Temporal Cue:} \emph{“In 1999, Lionel Messi was a member of sports team ...”}
\end{itemize}

By inserting noise (or other forms of corruption) into specific tokens (often the subject or the year token) and selectively restoring only certain hidden states, we measure how each portion of the input affects final predictions. 
Our experiments on a Llama2 model highlight that \emph{subject tokens}, when combined with a year, incur the largest impact on retrieving the correct year-specific fact.

\subsection{Year-Based Causal Tracing of Subject Tokens}

\paragraph{Heatmap Illustrations}
The top 6 plots in Figures~\ref{fig:causal_tracing} depict example heatmaps for \emph{single-layer} restoration (left) vs.\ \emph{MLP-interval} and \emph{Attention-interval} restoration (center, right). 
Each subplot visualizes how restoring a given layer (or set of layers) changes the probability of a target answer (e.g., \(\mathrm{p}(\text{New})\) or \(\mathrm{p}(\text{Barcelona})\)). 
Darker regions indicate larger improvements in the model’s correctness after that restoration. 
We compare:

\begin{itemize}
    \item \textbf{Top row:} Restoration effect on \(\mathrm{p}(\text{New})\) or \(\mathrm{p}(\text{Barcelona})\) for different single or grouped layers, showing which layers are most responsible for \emph{selecting} a new or correct team.
    \item \textbf{Bottom row:} Similar restoration but for alternative completions (e.g., \(\mathrm{p}(2)\) or \(\mathrm{p}(\text{Lion})\)), revealing how subject or year tokens can shift the model’s internal preference.
\end{itemize}

We observe that certain mid-range layers, especially around 10--20, exhibit strong spikes: 
when we restore those layers’ subject-year hidden states, the model reverts to a correct or plausible answer for the year-specific query.

\paragraph{Time Affects the \emph{Subject} Most}
As hinted by the heatmaps:
\begin{itemize}
    \item The \emph{largest gain in correct probability} typically occurs after restoring subject+year hidden states. 
    If corrupted, the model confuses or misaligns the year with the wrong subject, yielding off-target outputs (e.g.\ a different team or a random hallucination).
    \item Other tokens (relation or object) produce \emph{smaller} jumps when restored. Although they matter for the final fact, they do not exhibit the same \emph{temporal} sensitivity as the subject domain.
\end{itemize}

\subsection{Year-Based Causal Tracing of Relation and Object Tokens}
\begin{figure*}[t]
\vspace{-20pt}
\begin{center}
    \includegraphics[width=1\textwidth]{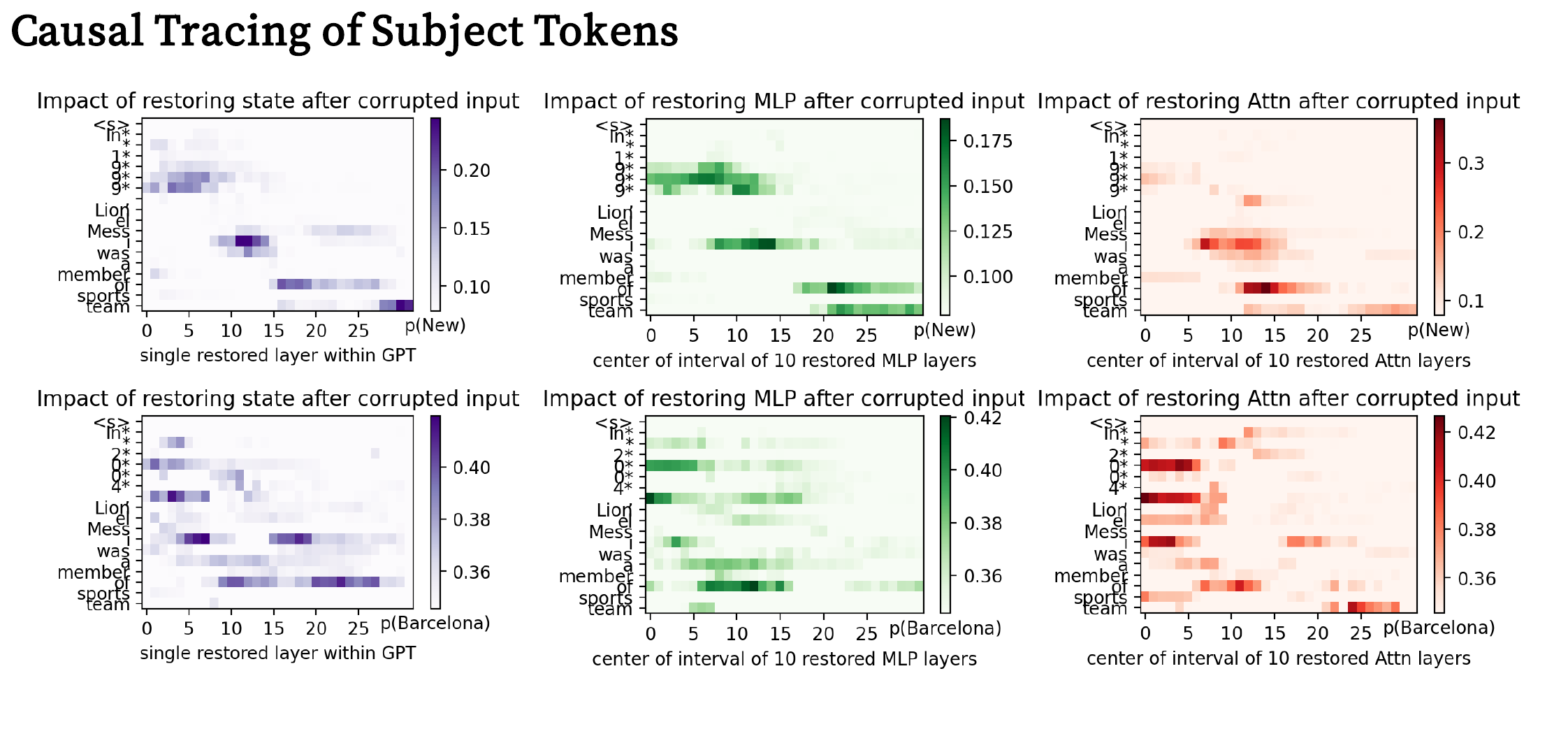}
    \vspace{-20pt}
    \includegraphics[width=1\textwidth]{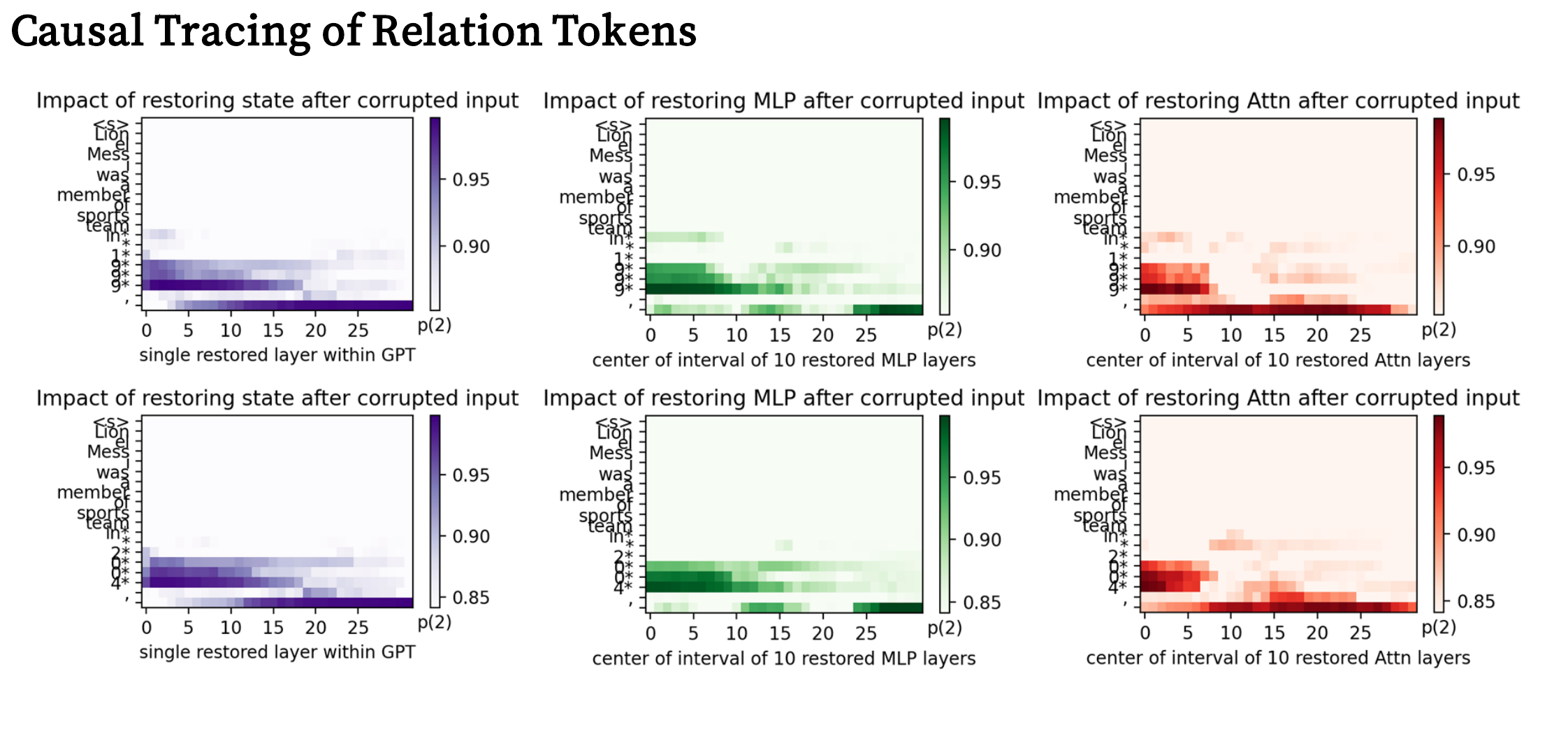}
    \includegraphics[width=1\textwidth]{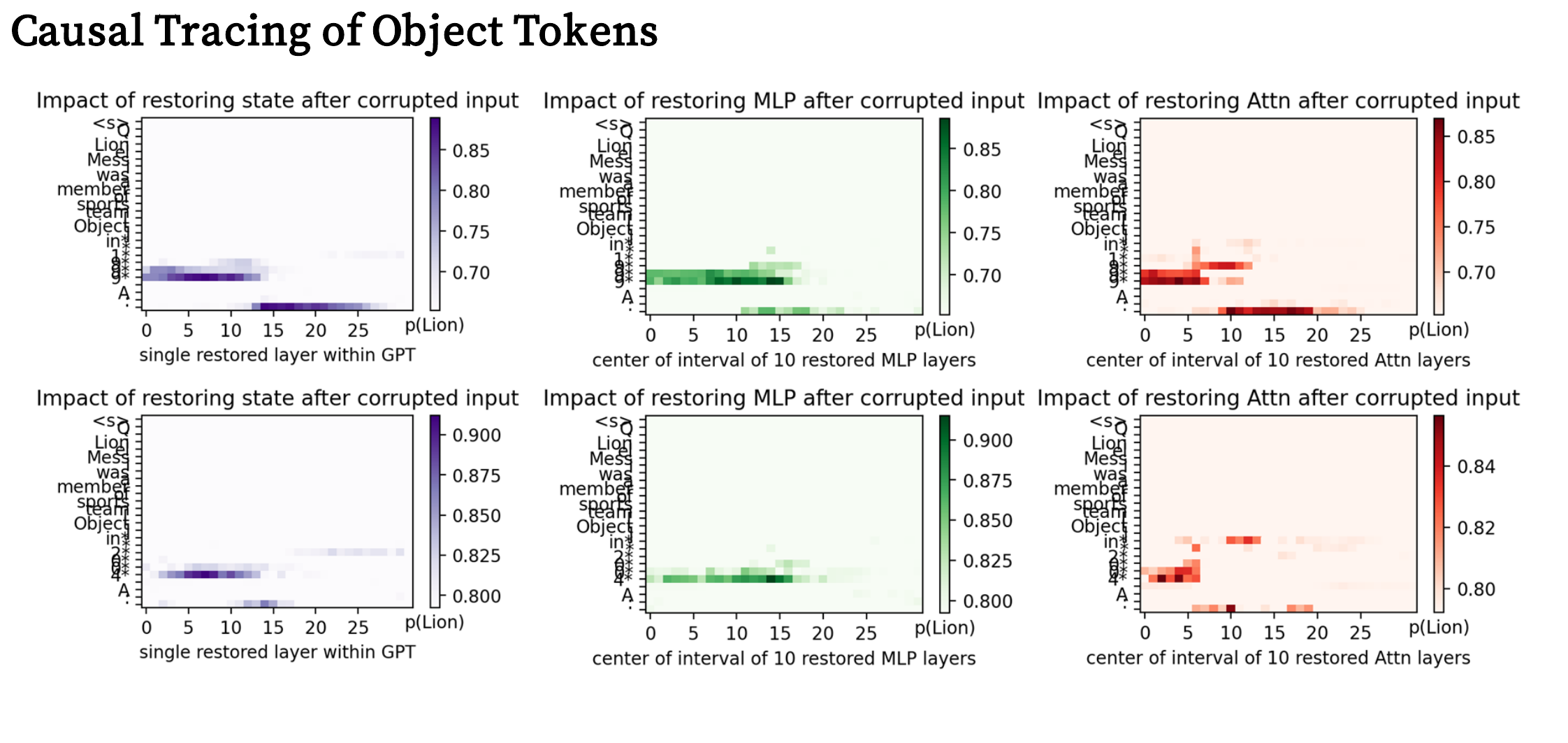}
\end{center}%
\vspace{-20pt}%
\caption{Results of Causal Tracing for all position(subject, relation, object), six plots for each cases from the top to middle and bottom. 
The restoring part is set to each temporal conditioning, in two different age: 1999 and 2004. 
(Illustrative) Causal tracing heatmaps showing how restoring different layers (x-axis) after temporal corruption affects $\mathrm{p}(\text{New})$ or $\mathrm{p}(\text{Barcelona})$. 
For the object position, we set a simulated \emph{[Object]} for the place holder.
Each figure's left column represents single-layer restoration; the center and right columns reflect MLP vs.\ attention intervals. 
Restoring subject+year at mid layers yields pronounced differences (dark regions).
On the other hand, restoring relation+year or object+year yields trivial differences as their range is overlap significantly.
}
\label{fig:causal_tracing}
\vspace{-10pt}
\end{figure*}

The middle and lower side six plots in Figures~\ref{fig:causal_tracing} replicate the above procedure for \emph{relation tokens} (e.g., “was a member of”) and \emph{object tokens} (e.g., a team name). 
The heatmaps show weaker or narrower restoration effects when the year corruption is placed near those tokens:

\begin{itemize}
    \item \textbf{Relation tokens} only yield modest probability recovery upon restoration, implying that while they shape the factual link, they do not anchor the \emph{time} dimension.
    \item \textbf{Object tokens} affect final correctness but appear less coupled to the year. Overwriting their hidden states helps for precise object naming, yet does not fix \emph{when} an event is said to occur.
\end{itemize}

\subsection{Implications for Temporal-Subject Coupling}
In line with prior studies~\cite{rome}, these findings confirm that the \emph{temporal aspect} is mainly fused into the \emph{subject} representation---the model effectively treats “(Subject in Year)” as a unique entity. 
Restoring the subject+year region of hidden states yields the greatest improvement, implying that year tokens attach strongly to the subject slot. 
Conversely, \emph{relation} and \emph{object} tokens are comparatively less sensitive to time cues.

\paragraph{Limitations of Causal Tracing Alone}
Despite highlighting \emph{which layer} or \emph{token positions} matter, causal tracing alone cannot pinpoint \emph{which heads or MLPs} form the circuit that routes these time signals. 
For instance, a single layer might have multiple attention heads with different behaviors; or an MLP might selectively process the year dimension but remain obscure at the token-level. 
As we explore in (\S\ref{sec:knw-circuit-reuse}), adopting a \emph{circuit-level} perspective unveils specific \emph{Temporal Heads} that systematically propagate year-conditioned knowledge throughout the model.

\begin{table}[t]
\centering
\vspace{-5pt}
{\resizebox{\columnwidth}{!}{
\begin{tabular}{llllll}
\toprule
\multicolumn{2}{l}{\textbf{Category}} & \textbf{Knowledge} & \textbf{\#Node} & \textbf{\#Edge} & \textbf{CRS} \\ 
\midrule
\multicolumn{6}{l}{\textbf{\textit{Temporal}}} \\ 
\midrule
Sports     &            & Nicolas Anelka     & 27  & 26  & 88.81 \\ 
    &            & David Beckham       & 42  & 59  & 26.50 \\ 
Presidents &            & Argentina          & 38  & 64 & 43.99 \\ 
 &            & South Korea        & 51  & 104 & 53.18 \\ 
CEO        &            & Hewlett-Packard    & 31  & 34 & 40.36 \\ 
       &            & Chrysler           & 26  & 22  & 28.14 \\ 
Defense    &            & United States      & 8  & 5 & 25.60 \\ 
 &            & China              & 13  & 9 & 25.82 \\ 
\midrule
\multicolumn{3}{l}{\textbf{Avg}} & \textbf{30} & \textbf{40} & \textbf{41.44} \\ 
\midrule
\multicolumn{6}{l}{\textbf{\textit{Time-Invariant}}} \\ 
\midrule
CommonSense             &            & Object Superclass  & 72  & 127  & 42.61 \\ 
Conditional CS &            & Fruit Inside Color & 43  & 49 & 64.83 \\ 
Num in Obj     &            & Geometric Shape    & 60  & 127 & 62.94 \\ 
Num in Sub     &            & Roman Numerals     & 57  & 108 & 71.18 \\ 
\midrule
\multicolumn{3}{l}{\textbf{Avg}} & \textbf{58} & \textbf{103} & \textbf{60.39} \\ 
\bottomrule
\end{tabular}}}{}
\caption{Statistics of temporal knowledge circuits for Qwen 1.5, both temporal and time-invariant knowledge.
For temporal knowledge, each type of knowledge is reproduced with three selected years: \textbf{1999, 2004, and 2009}.
The numbers of nodes, edges and CRS is the average of each knowledge's yearly circuits.
We simplified total circuits with $\tau = 0.1$, same as Llama2.
}
\label{table:statistic_crs_qwen}
\vspace{-10pt}
\end{table}

\section{Details of Circuit Reproduction Score}
\label{sec:detail_in_crs}
CRS condenses relative performance differences and sign alignment into a single, intuitive 0–100 metric, offering a streamlined assessment of circuit quality.

\subsection{Motivation}
Existing approaches such as \textit{logit diff} or \textit{MatchNLL}~\citep{conmy2023towards, KC} evaluate circuits by reporting two separate numbers: the \textbf{baseline performance} of the original model and the \textbf{circuit’s performance}. However, this can obscure direct comparisons, especially when values are of different scales or signs. To address this, we introduce the \textbf{Circuit Reproduction Score (CRS)}, a unified metric that normalizes these comparisons onto a \textbf{0–100 scale}. A score of 0 indicates a circuit that fails to retain meaningful model behavior, while 100 signifies equal or superior performance compared to the original model.

\subsection{Definition}
Let $B$ represent the \textbf{baseline performance} of the original model and $P$ the \textbf{circuit’s performance}. CRS is computed as:
\begin{equation}
CRS(B, P) = 100 \times S(B,P) \times D(B,P),
\end{equation}
where:
\begin{itemize}
    \item $S(B,P) \in (0,1]$ is a sign-based adjustment factor.
    \item $D(B,P) = \exp(-\alpha R)$ scales based on deviation $R$.
    \item $\alpha$ controls the sensitivity to deviations.
\end{itemize}

The deviation $R$ is defined as:
\begin{equation}
R = \frac{\text{dist}(B,P)}{|B|},
\end{equation}
where $\text{dist}(B,P)$ measures how far $P$ deviates from $B$.

If the circuit’s performance meets or exceeds the baseline ($B > 0$ and $P \geq B$), CRS is set to:
\begin{equation}
CRS(B, P) = 100.
\end{equation}

\begin{table}[t]
\centering
\vspace{-5pt}
{\resizebox{\columnwidth}{!}{
\begin{tabular}{llllll}
\toprule
\multicolumn{2}{l}{\textbf{Category}} & \textbf{Knowledge} & \textbf{\#Node} & \textbf{\#Edge} & \textbf{CRS} \\ 
\midrule
\multicolumn{6}{l}{\textbf{\textit{Temporal}}} \\ 
\midrule
Sports     &            & Nicolas Anelka     & 5  & 3  & 64.51 \\ 
    &            & David Beckham       & 22  & 22  & 42.24 \\ 
Presidents &            & Argentina          & 53  & 127 & 91.19 \\ 
 &            & South Korea        & 55  & 142 & 81.47 \\ 
CEO        &            & Hewlett-Packard    & 12  & 9 & 35.55 \\ 
       &            & Chrysler           & 9  & 7  & 73.98 \\ 
Defense    &            & United States*      & 3  & 1 & 73.03 \\ 
 &            & China*              & 2  & 1 & 72.85 \\ 
\midrule
\multicolumn{3}{l}{\textbf{Avg}} & \textbf{20} & \textbf{39} & \textbf{66.85} \\ 
\midrule
\multicolumn{6}{l}{\textbf{\textit{Time-Invariant}}} \\ 
\midrule
CommonSense             &            & Object Superclass  & 73  & 135  & 61.49 \\ 
Conditional CS &            & Fruit Inside Color & 24  & 44 & 49.48 \\ 
Num in Obj     &            & Geometric Shape    & 16  & 20 & 39.98 \\ 
Num in Sub     &            & Roman Numerals     & 78  & 153 & 74.04 \\ 
\midrule
\multicolumn{3}{l}{\textbf{Avg}} & \textbf{48} & \textbf{88} & \textbf{56.25} \\ 
\bottomrule
\end{tabular}}}{}
\caption{Statistics of temporal knowledge circuits for Phi 3 mini, both temporal and time-invariant knowledge.
For temporal knowledge, each type of knowledge is reproduced with three selected years: \textbf{1999, 2004, and 2009}.
The numbers of nodes, edges and CRS is the average of each knowledge's yearly circuits.
We simplified total circuits with $\tau = 0.1$, same as Llama2, except knowledge in Defense where at least 30\% lower $\tau$ is needed.
Interestingly, Phi 3 mini shows better CRS of temporal knowledge than time-invariant ones, though their overall simplified nodes and edges are less than same cases of other models.
}
\label{table:statistic_crs_phi}
\vspace{-10pt}
\end{table}

\begin{table}[t]
\centering
\resizebox{\columnwidth}{!}{
\begin{tabular}{lll}
\toprule
\textbf{Category} & \textbf{Time Range} & \textbf{\# of Cases} \\ 
\midrule
\multicolumn{3}{l}{\textbf{\textit{Temporal Knowledge~\citep{wikidata}}}} \\ 
\midrule
Sports             & 1996-2020  & 81  \\ 
Presidents         & 1999-2009  & 65  \\ 
CEO                & 1999-2009  & 65  \\ 
Defense            & 1999-2009  & 77  \\ 
Movies             & 1999-2009  & 33  \\ 
GDP                & 1999-2009  & 33  \\ 
Inflations         & 1999-2009  & 33  \\ 
\midrule
\multicolumn{3}{l}{\textbf{\textit{Time Invariant Knowledge~\citep{lre}}}} \\ 
\midrule
Object Superclass  & -          & 36  \\ 
Fruit Inside Color & -          & 76  \\ 
Geometric Shape    & -          & 28  \\ 
Roman Numerals     & -          & 31  \\ 
\bottomrule
\end{tabular}
}
\caption{Statistics of dataset used for circuits.
}
\label{table:statistic_dataset1}
\end{table}

\begin{table}[t]
\centering
\resizebox{\columnwidth}{!}{
\begin{tabular}{llll}
\toprule
\textbf{Dataset}            & \textbf{Format} & \textbf{Test} & \textbf{Source} \\
\midrule
TriviaQA           & MCQA   & 11,313  & \citealp{triviaqa}         \\
Math ChroKnowledge & MCQA   & 2,585  & \citealp{mathkg, chroknowledge}          \\
\bottomrule
\end{tabular}
}
\caption{Statistics of dataset used general QA.
}
\label{table:statistic_dataset2}
\end{table}

\paragraph{Handling Positive and Negative Baselines}
\begin{itemize}
    \item If $B > 0$ and $P \geq B$, CRS is 100, indicating that the circuit fully retains or improves upon original performance.
    \item If $P < B$, the CRS score is exponentially reduced based on the relative performance gap.
    \item If $B < 0$ (indicating the original model performed poorly), less negative performance is treated as an improvement.
    \item If $B$ and $P$ differ in sign, CRS applies an intermediate weighting (e.g., 0.6–0.8) to avoid misleadingly high scores.
\end{itemize}

\subsection{Implementation}
We compute:
\begin{align}
B &= \text{eval\_baseline}(G, \mathcal{D}_{val}, \text{logit\_diff}), \\
P &= \text{eval\_graph}(G, P, \mathcal{D}_{val}, \text{logit\_diff}).
\end{align}
These yield average performance values, which are then converted into:
\begin{equation}
CRS = \text{one\_score}(B, P; \alpha, S) \in [0,100].
\end{equation}

The resulting CRS provides a concise and interpretable measure of circuit faithfulness:
\begin{itemize}
    \item \textbf{Both negative:} The circuit’s score is capped (e.g., at most $100 \times 0.5$).
    \item \textbf{Both positive:} The circuit may reach 100 if it fully retains baseline performance.
    \item \textbf{Mixed sign:} An intermediate factor (e.g., 0.6–0.8) prevents inflated scores if the circuit behaves in an unintended manner.
\end{itemize}

\subsection{Hyperparameters}
The CRS computation relies on several hyperparameters that modulate its sensitivity to deviations and its handling of different sign scenarios:
\begin{itemize}
    \item $\alpha$: Sensitivity to deviation, controlling how sharply CRS decreases as the circuit deviates from the baseline. Default: 1.0.
    \item $sf_{\text{bothpos}}$: Sign factor when both baseline and circuit performance are positive ($B > 0$, $P > 0$). Default: 1.0.
    \item $sf_{\text{bothneg}}$: Sign factor when both baseline and circuit performance are negative ($B < 0$, $P < 0$). Default: 0.5.
    \item $sf_{\text{bneg\_cpos}}$: Sign factor when the baseline is negative but the circuit is positive ($B < 0$, $P > 0$). Default: 0.8.
    \item $sf_{\text{bpos\_cneg}}$: Sign factor when the baseline is positive but the circuit is negative ($B > 0$, $P < 0$). Default: 0.6.
    \item $\epsilon$: Small constant for numerical stability, ensuring nonzero denominators and preventing division errors. Default: $10^{-9}$.
\end{itemize}

\section{Details and Statistics of Dataset}
\label{sec:dataset_details}
Table~\ref{table:statistic_dataset1} and~\ref{table:statistic_dataset2} present the statistical details of the knowledge datasets used in our evaluation. 
For temporal knowledge, we utilize open-sourced WikiData as referenced. 
These datasets encompass a variety of knowledge categories, each consisting of multiple objects along with their associated time ranges.

\subsection{Categorization of Knowledge Datasets}
Each dataset category represents a specific type of structured knowledge:

\paragraph{Temporal Knowledge.} This category contains knowledge that varies over time, requiring temporal awareness for accurate retrieval. 
The definitions for each subcategory are as follows:
\begin{itemize}
    \item \textbf{Sports}: The teams associated with specific athletes over time.
    \item \textbf{Presidents}: The names of country leaders for given years.
    \item \textbf{CEO}: The chief executive officers of major companies in a given year.
    \item \textbf{Defense}: The national defense budget of different countries.
    \item \textbf{Movies}: The highest-grossing films by country for specific years.
    \item \textbf{GDP}: The annual Gross Domestic Product (GDP) of various countries.
    \item \textbf{Inflation}: The inflation rate of different countries for given years.
\end{itemize}

\begin{table*}[]
\centering
{\scalebox{0.9}{%
\begin{tabular}{llllllllll}
\toprule
\multicolumn{1}{c}{\multirow{2}{*}{\textbf{Settings}}} & \multicolumn{7}{c}{\textbf{Temporal Knowledge (\%)}} & \multicolumn{1}{c}{\multirow{2}{*}{\textbf{Average}}} \\ \cmidrule(lr){2-8}
\multicolumn{1}{c}{} & \multicolumn{1}{c}{\textbf{Sports}} & \multicolumn{1}{c}{\textbf{Presidents}} & \multicolumn{1}{c}{\textbf{CEO}} & \multicolumn{1}{c}{\textbf{Defense}} & \multicolumn{1}{c}{\textbf{Movies}} & \multicolumn{1}{c}{\textbf{GDP}} & \multicolumn{1}{c}{\textbf{Inflations}} & \multicolumn{1}{c}{} \\ \midrule
\rowcolor[HTML]{BFD9EC} 
\multicolumn{9}{c}{\textbf{\textit{Llama-2-7b-chat-hf - a18,h3, a15.h0}}} \\ \midrule
Baseline & 41.9 & 80.7 & 27.5 & 13.5 & 23.1 & 10.4 & 10.8 & 29.7 \\
Ablation & \textcolor{red}{40.0} & 75.6 & \textcolor{red}{21.3} & 13.3 & \textcolor{red}{9.37} & 10.7 & 9.34 & 25.6 \\ \midrule
\rowcolor[HTML]{D4BFE1} 
\multicolumn{9}{c}{\textbf{\textit{Qwen1.5-7B-Chat - a17.h15}}} \\ \midrule
Baseline & 32.4 & 57.2 & 19.6 & 11.5 & 16.7 & 9.58 & 10.0 & 22.4 \\
Ablation & 32.0 & \textcolor{red}{49.4} & 16.6 & 10.3 & 10.8 & 9.50 & 10.3 & 19.8 \\ \midrule
\rowcolor[HTML]{8EDB8A}
\multicolumn{9}{c}{\textbf{\textit{Phi-3-mini-4k-instruct - a10.h13}}} \\ \midrule
Baseline & 24.4 & 72.1 & 30.8 & 73.7 & 21.4 & 12.2 & 13.5 & 35.4 \\
Ablation & 24.8 & 69.6 & 30.7 & \textcolor{red}{11.5} & 21.6 & \textcolor{red}{11.7} & \textcolor{red}{11.8} & \textcolor{red}{26.0} \\ \bottomrule
\end{tabular}}}{}
\caption{Total results of temporal knowledge across multiple models.
Each scores were measured in probability (\%) with averaging effect of multiple heads ablation results.
The most dropped score for each column is colored red.}
\label{table:total_result1}
\end{table*}

\paragraph{Time-Invariant Knowledge.} Unlike temporal knowledge, this category consists of facts that do not change over time.
The specific subcategories are defined as follows:
\begin{itemize}
    \item \textbf{Object Superclass}: General commonsense knowledge that categorizes objects into broader superclasses.
    \item \textbf{Fruit Inside Color}: Commonsense knowledge conditioned on the phrase ``On the inside,'' focusing on the internal color of fruits.
    \item \textbf{Geometric Shape}: Knowledge where objects are associated with numerical properties, such as shape classifications based on the presence of numbers.
    \item \textbf{Roman Numerals}: Cases where numerical values appear in the subject itself, typically involving Roman numeral representations.
\end{itemize}

\subsection{General Question Answering (QA) Datasets}
In addition to the structured knowledge datasets, we also utilize benchmark QA datasets for evaluation. 
The test or validation sets provided by these benchmarks are used in our experiments. 
All evaluations are conducted under the \textbf{Multiple-Choice Question Answering (MCQA)} setting.
Statistics are following Table~\ref{table:statistic_dataset2}.

\section{Details of Log Probability Check}
\label{sec:log_evaluation_details}
Our evaluation follows the paradigm outlined in \citealt{logprob}, focusing on log probability variation rather than direct answer accuracy.  
Standard multiple-choice evaluations often overestimate model difficulty by testing answers in isolation rather than in comparative contexts.
Instead, we analyze how ablation affects probability distributions across all candidate objects, providing a more granular view of temporal knowledge representation.
By using per-object probability tracking, we reveal a more precise representation of how temporal information is encoded and manipulated within the model.
\paragraph{Notations}
Let \( M \) be the transformer model under evaluation, and let \( O \) be the set of all candidate objects (e.g., teams, presidents).  
For a given input, the model assigns a probability \( p(o | s,r,T) \) to each object \( o \in O \) with given subject $s$, relation $r$ and time $T$, representing its likelihood of being the correct answer.  
The object assigned the highest probability is labeled \texttt{Target} if it corresponds to the correct temporal fact, or \texttt{Non-Target} otherwise.
\paragraph{Per-Choice Probability Assessment}
Unlike conventional approaches, which focus solely on the final prediction, we track probability variations across all objects.  
This ensures that we capture nuanced knowledge shifts caused by ablation, rather than just observing whether the top-ranked answer changes.

\begin{table*}[]
\centering
{\scalebox{0.8}{%
\begin{tabular}{llllllll}
\toprule
\multicolumn{1}{c}{\multirow{2}{*}{\textbf{Settings}}} & \multicolumn{5}{c}{\textbf{Time Invariant Knowledge (\%)}} & \multicolumn{2}{c}{\textbf{General QA (F1 \& \%)}} \\ \cmidrule(lr){2-6} \cmidrule(lr){7-8}
\multicolumn{1}{c}{} & \multicolumn{1}{c}{\textbf{Obj-Super}} & \multicolumn{1}{c}{\textbf{Fruit In-Color}} & \multicolumn{1}{c}{\textbf{Geo-Shape}} & \multicolumn{1}{c}{\textbf{Roman-Num}} & \multicolumn{1}{c}{\textbf{Average}} & \multicolumn{1}{c}{\textbf{TriviaQA}} & \multicolumn{1}{c}{\textbf{Math}} \\ \midrule
\rowcolor[HTML]{BFD9EC} 
\multicolumn{8}{c}{\textbf{\textit{Llama-2-7b-chat-hf - a18,h3, a15.h0}}} \\ \midrule
Baseline & 49.7 & 75.6 & 68.5 & 53.5 & 61.8 & 55.4 & 45.4 \\
Ablation & 50.2 & 75.6 & 68.1 & 53.0 & 61.7 & 54.9 & 45.3 \\ \midrule
\rowcolor[HTML]{D4BFE1} 
\multicolumn{8}{c}{\textbf{\textit{Qwen1.5-7B-Chat - a17.h15}}} \\ \midrule
Baseline & 48.0 & 72.0 & 69.4 & 61.5 & 62.7 & 49.7 & 77.0 \\
Ablation & 47.8 & 72.0 & 69.3 & 61.1 & 62.6 & 49.5 & 77.0 \\ \midrule
\rowcolor[HTML]{8EDB8A}
\multicolumn{8}{c}{\textbf{\textit{Phi-3-mini-4k-instruct - a10.h13}}} \\ \midrule
Baseline & 21.8 & 76.0 & 68.3 & 73.2 & 59.8 & 46.8 & 80.8 \\
Ablation & 23.2 & 76.4 & 69.1 & 73.7 & 60.6 & 46.2 & 81.2 \\ \bottomrule
\end{tabular}}}{}
\caption{Total results of time invariant knowledge and general QA across multiple models.
For TriviaQA, we test the unfiltered, no-context validation set (11.3k).
Each scores were measured in probability (\%) or f1 score with averaging effect of multiple heads ablation results.
Most of cases, the scores remain stable or even goes up such as \emph{Object Superposition}.}
\label{table:total_result2}
\end{table*}

\paragraph{Head Ablation and Probability Recalculation}
To examine the role of temporal attention heads, we zero out selected heads \( \hat{H} \) and measure how the model's probability distribution over \( O \) shifts.  
The recalculated probability after ablation is given by:
\begin{align}
    z_o &= \log p_\text{ablate}(o|s,r,T),
    \\ \hat{p}_o &= \frac{\exp(z_o)}{\sum_{o' \in O} \exp(z_{o'})},
\end{align}
where \( p_{\text{ablate}} \) denotes the log-probability computed by forward pass of model, with ablation of corresponding heads in \( \hat{H} \).
Unlike standard evaluation, this method isolates the impact of specific attention heads on temporal knowledge retention.

\section{Total Result Each Datasets}
\label{app:total_qa}
Table~\ref{table:total_result1}--\ref{table:total_result2} indicates total result of time variant, invariant and general QA for all three models.
We additionally deal with the case of Movies (which movie is the most popular in each year for each countries), GDP (how much GDP for each year for each countries) and Inflation (the inflation rate of each countries).
As colored in red, temporal knowledge drops more drastically than time invariant knowledge or general QA.

\section{Details of Temporal Knowledge Editing}
\label{appendix:temp-edit}

\subsection{Attention Value Extraction and Injection}
We employ a direct attention value addition method to influence the model’s temporal knowledge representation.
Though we inspired by activation addition or patching methods like ~\citealp{actaddllama2, iti, cast, saevector} and especially \citealp{actadd}, which computes an activation difference between positive and negative prompts, our method directly extracts value of attention heads from the \texttt{source\_prompt} and injects them into the \texttt{target\_prompt}.

\paragraph{Extracting Value of Attention Head}
For a given \texttt{source\_prompt}, we extract the value from a specific attention head \((l,h)\) at the token position corresponding to the temporal entity:
\begin{equation}
    \mathbf{a}_{\mathrm{src}} = \text{AttnV}(x_{\text{src}}, l, h),
\end{equation}
where \( x_{\text{src}} \) is the tokenized \texttt{source\_prompt} and \( \text{AttnV}(x, l, h) \) returns the attention value at layer \( l \) and head \( h \).

To obtain a stable representation across multiple \texttt{source\_prompt}s, we compute the mean value:
\begin{equation}
    \mathbf{a}_{\mathrm{src}} = \frac{1}{N} \sum_{i=1}^{N} \text{AttnV}(x_{\text{src}}^{(i)}, l, h),
\end{equation}

\paragraph{Identifying Temporal Token Position}
In the \texttt{target\_prompt}, we locate the last token index of the temporal condition to determine where the \text{AttnV} injection should occur.

\paragraph{Attention Value Injection}
Once the temporal token index \( t_{\text{subj}} \) is found, we inject the extracted \text{AttnV}:
\begin{equation}
    \mathbf{a}_{\mathrm{tgt}} = \text{AttnV}(x_{\text{tgt}}, l, h),
\end{equation}
\begin{equation}
    \mathbf{a}_{\mathrm{tgt}}^{\text{new}} = \mathbf{a}_{\mathrm{tgt}} + \lambda \mathbf{a}_{\mathrm{src}},
\end{equation}
where \( x_{\text{tgt}} \) is the tokenized \texttt{target\_prompt}, \( \lambda \) is the injection coefficient (\(\lambda \in \{1, 3, 6\}\)), and \( \mathbf{a}_{\mathrm{tgt}}^{\text{new}} \) is the modified value.
This modification is applied dynamically using a forward hook:
\begin{equation}
    \text{Hook}(\mathbf{a}) = \mathbf{a} + \lambda \mathbf{a}_{\mathrm{src}}, 
\end{equation}
where $t = t_{\text{temp}}$ and \( x_{\text{temp}} \) is the tokenized temporal condition (e.g., "In 2009").

\subsection{Evaluation Metrics}
To assess the impact of attenion value injection, we introduce two evaluation criteria.

\paragraph{First-Token Prediction Shift}
We measure whether the injected value shifts the model’s predicted first token. 
Given the target prompt \( x_{\text{tgt}} \), we compare the probability of the correct response \( w^* \) before and after injection:
\begin{equation}
    P(w^* | x_{\text{tgt}}) < P(w^* | x_{\text{tgt}}^{\text{new}}),
\end{equation}
where \( P(w^* | x_{\text{tgt}}) \) is the original probability of generating the correct token and \( P(w^* | x_{\text{tgt}}^{\text{new}}) \) is the probability after attention value injection.

This probability shift is measured using log-probabilities from the model's output distribution.

\paragraph{Full-Text Response Validation}
To further verify the efficacy of our method, we check whether the model’s full generated response contains the expected factual entity. Specifically, we count the number of experiments where the correct answer appears in the model's output (e.g., "Dmitry Medvedev" for the name of president of Russia in 2009).

\begin{figure*}[t]
\vspace{-10pt}
\begin{center}
    \includegraphics[width=0.7\textwidth]{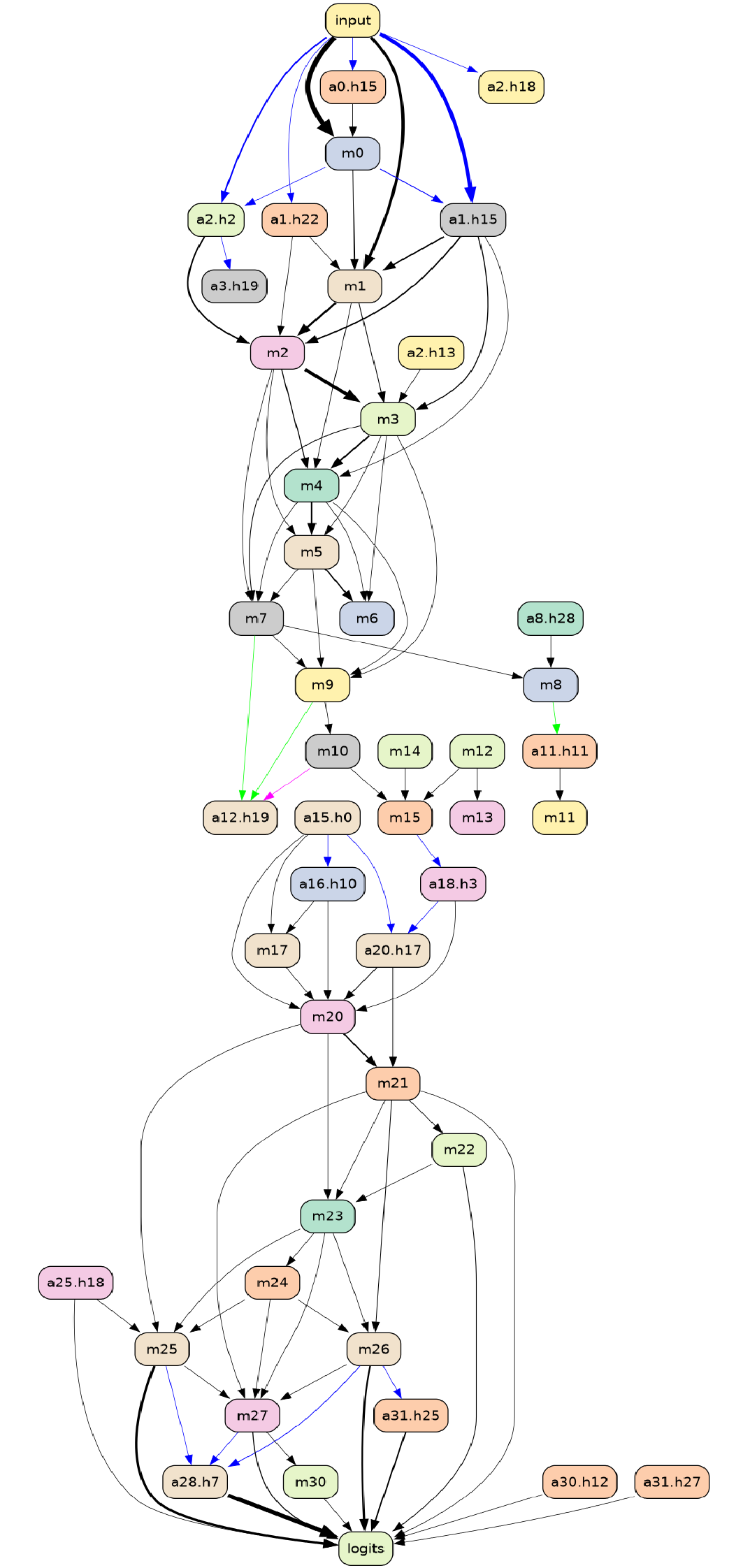}
\end{center}%
\vspace{-10pt}%
\caption{Temporal knowledge circuit of Llama2.
It is simplified version of total circuit by its importance of each nodes using $\tau = 0.1$ as threshold.
}
\label{fig:total_circuit}
\vspace{10pt}
\end{figure*}

\begin{figure*}[t]
\vspace{-10pt}
\begin{center}
    \includegraphics[width=0.8\textwidth]{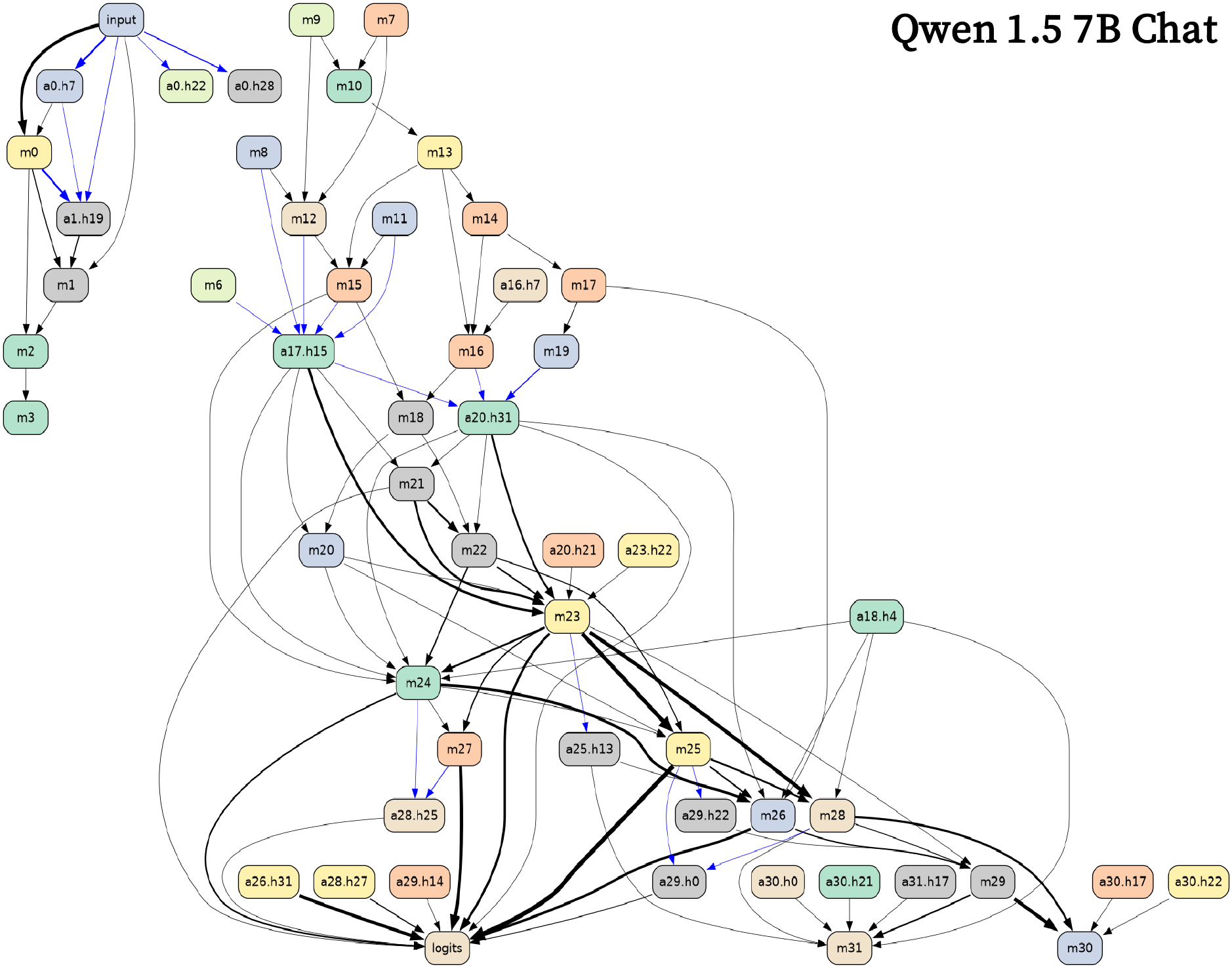}
\end{center}%
\vspace{20pt}
\begin{center}
    \includegraphics[width=0.8\textwidth]{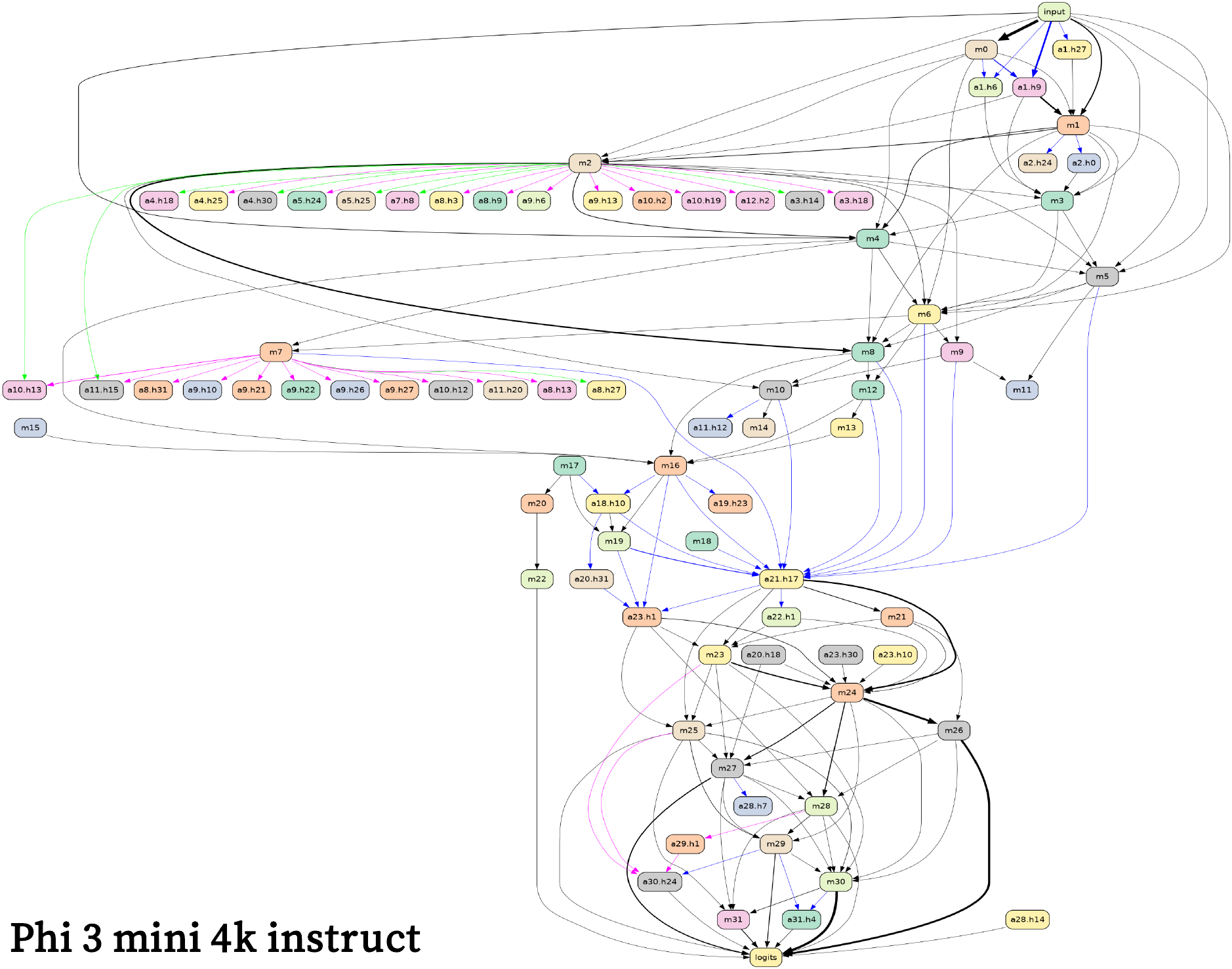}
\end{center}%
\vspace{10pt}%
\caption{Temporal knowledge circuit of Qwen 1.5 and Phi 3 mini.
Those are simplified version of total circuit according to each nodes and edges' importance of using same $\tau = 0.1$ as threshold.
}
\label{fig:total_circuit2}
\vspace{-10pt}
\end{figure*}

\begin{figure*}[t]
\vspace{-20pt}
\begin{center}
    \includegraphics[width=0.85\textwidth]{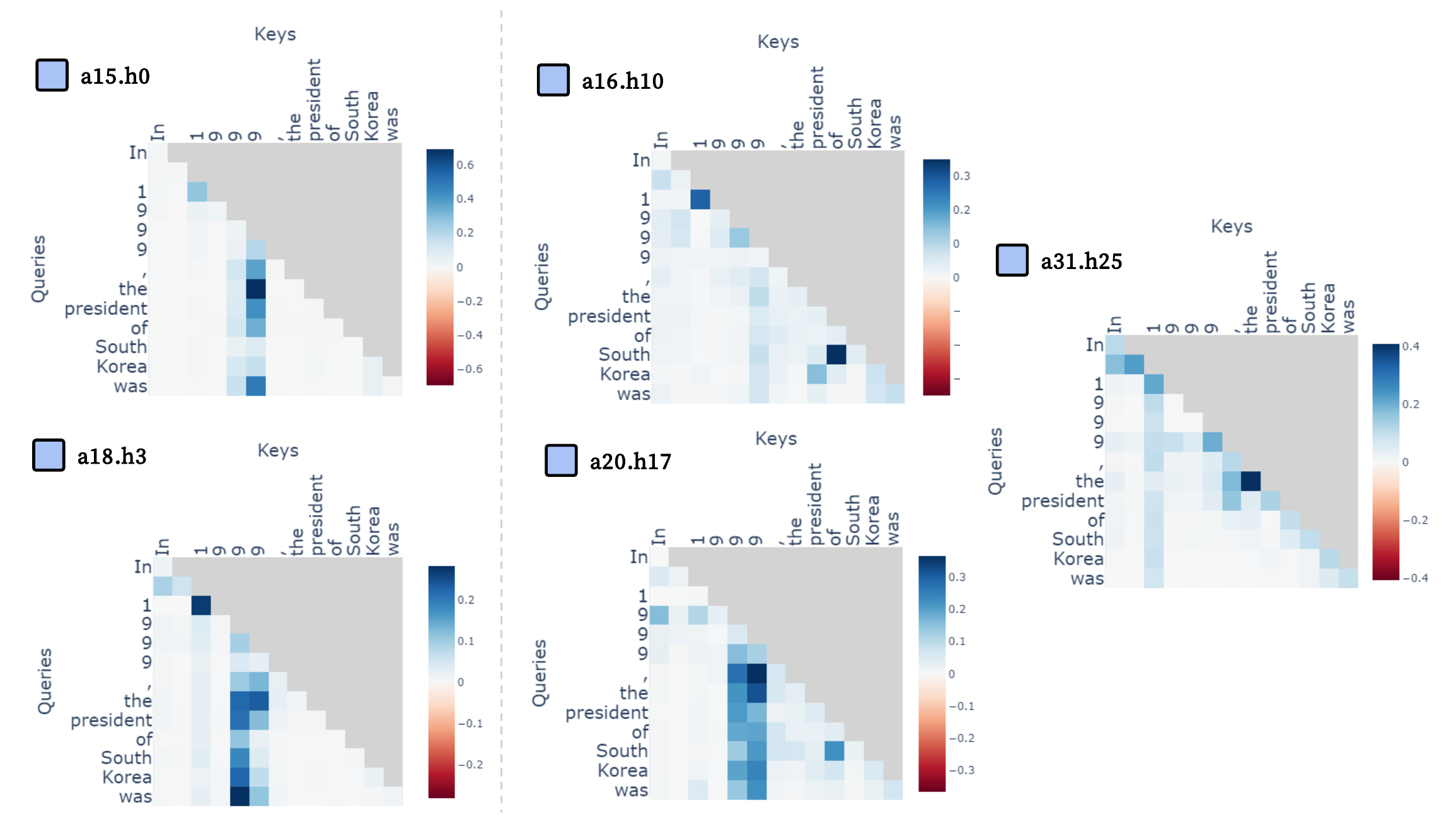}
\end{center}%
\vspace{-10pt}%
\caption{Total map of attention with Llama2-7b-chat-hf, for each temporal heads and backup temporal heads.
The left side of border line is the attention map of \textbf{Temporal Heads}, and the other side is the result of \textbf{Backup Temporal Heads}.
}
\label{fig:full_attn}
\vspace{-10pt}
\end{figure*}

\begin{figure*}[t]
\vspace{-10pt}
\begin{center}
    \includegraphics[width=0.85\textwidth, trim=0 80 0 80, clip]{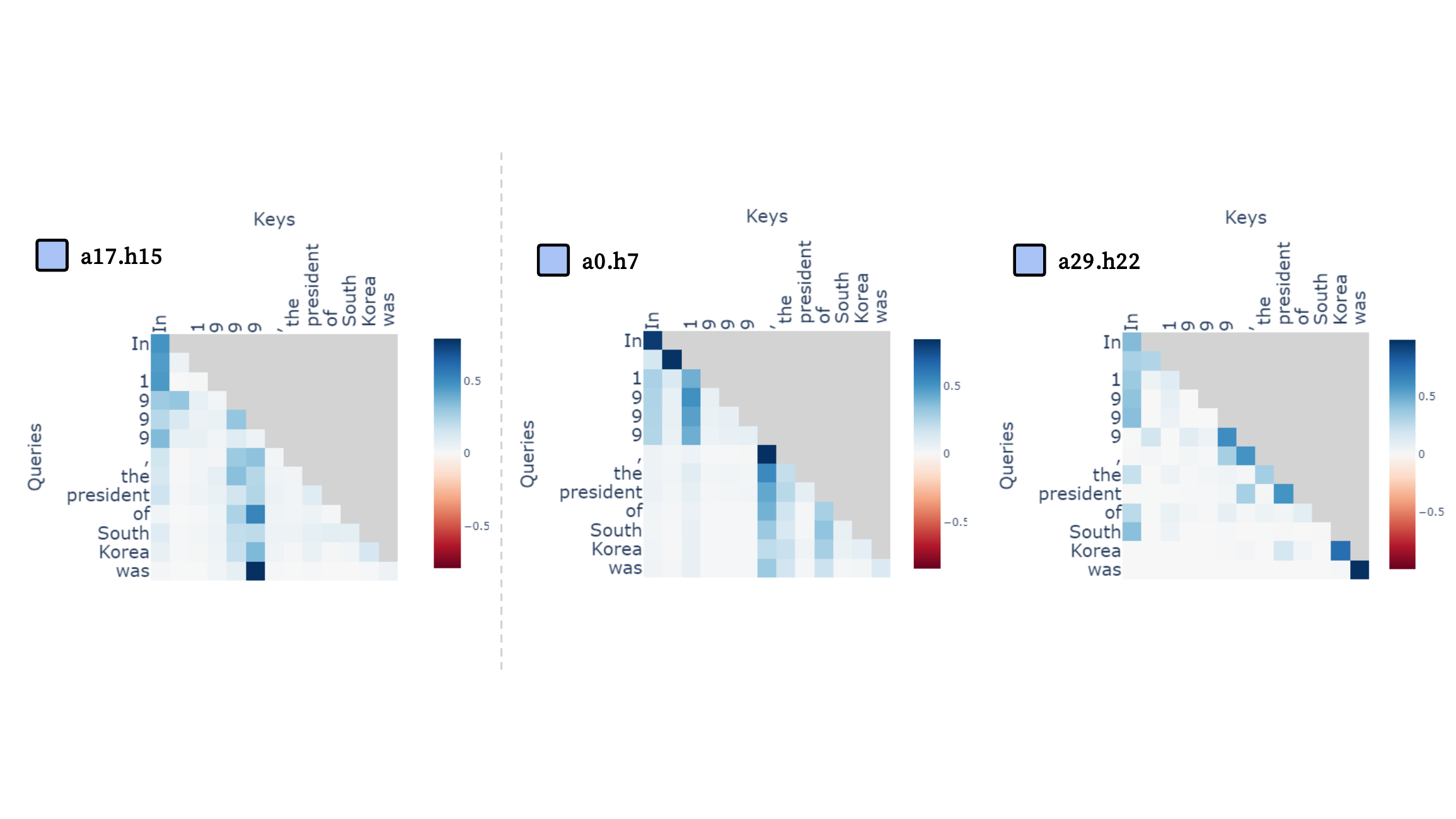}
\end{center}%
\vspace{-10pt}%
\caption{Total map of attention with Qwen1.5-7B-Chat, for each temporal heads and backup temporal heads.
The left side of border line is the attention map of \textbf{Temporal Heads}, and the other side is the result of \textbf{Backup Temporal Heads}.
}
\label{fig:full_attn_qwen}
\vspace{-10pt}
\end{figure*}

\begin{figure*}[t]
\vspace{-10pt}
\begin{center}
    \includegraphics[width=0.85\textwidth]{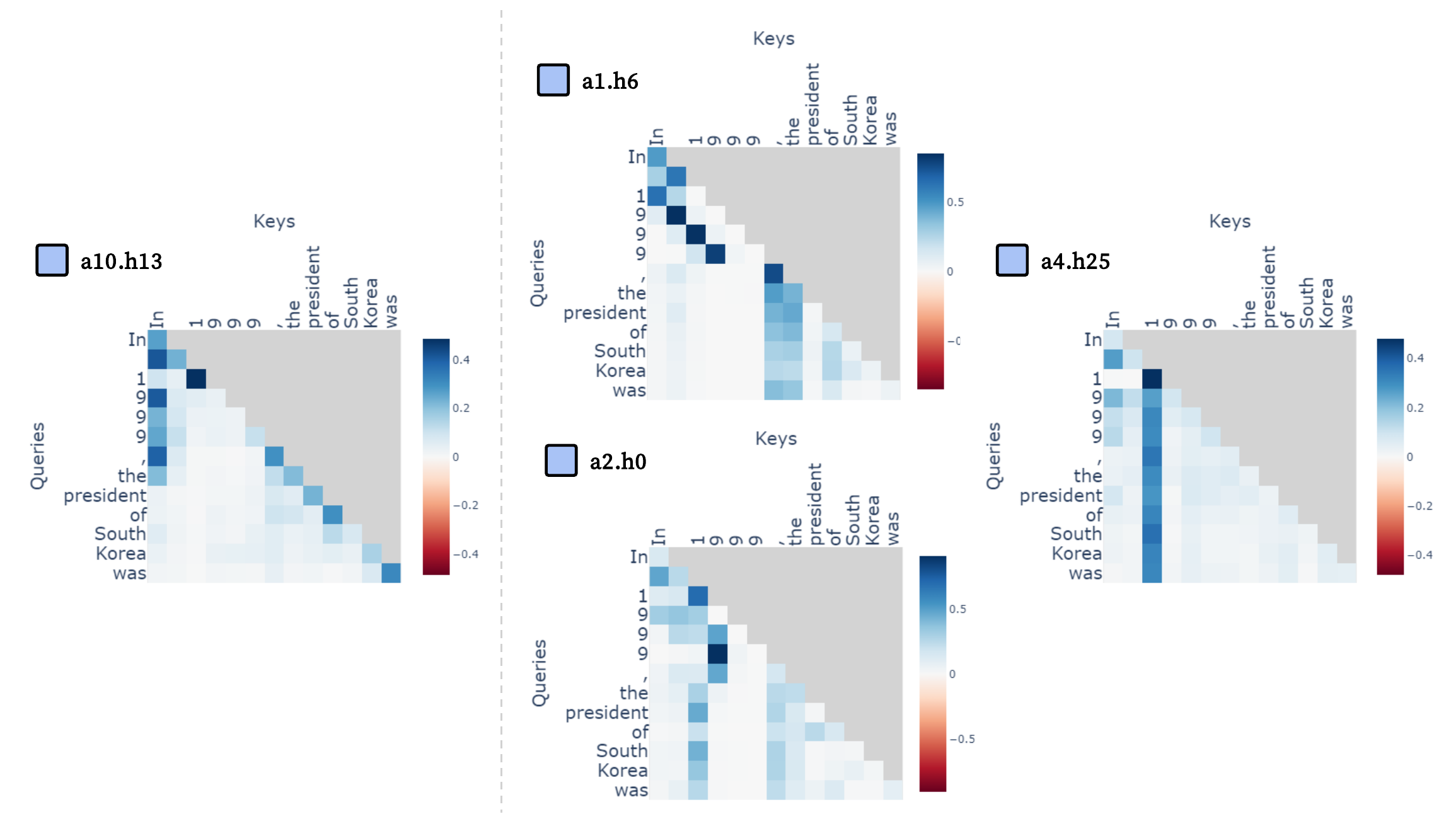}
\end{center}%
\vspace{-10pt}%
\caption{Total map of attention with Phi-3-mini-4k-instruct, for each temporal heads and backup temporal heads.
The left side of border line is the attention map of \textbf{Temporal Heads}, and the other side is the result of \textbf{Backup Temporal Heads}.
}
\label{fig:full_attn_phi}
\vspace{-10pt}
\end{figure*}

\begin{figure*}[t]
\vspace{-10pt}
\begin{center}
    \includegraphics[width=0.7\textwidth]{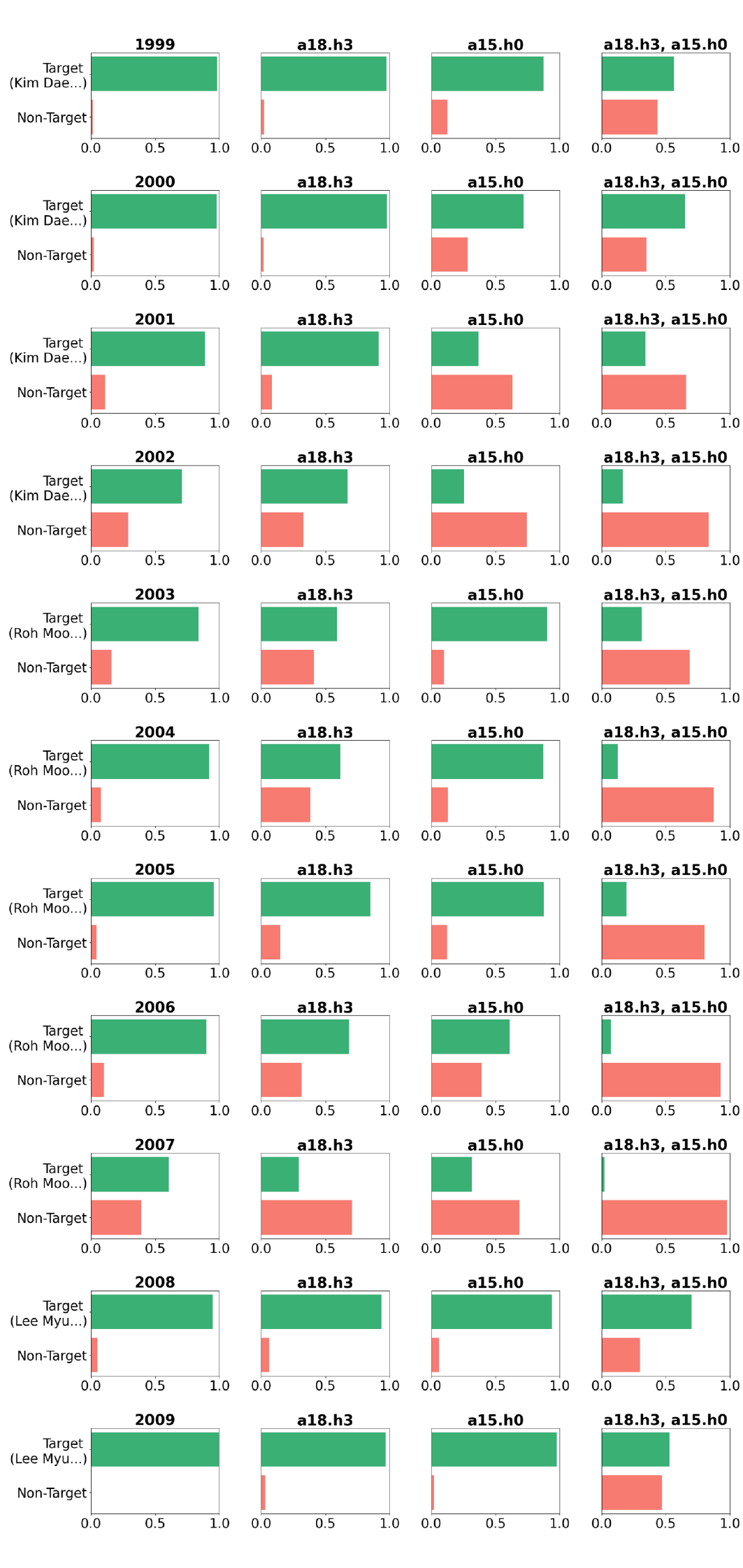}
\end{center}%
\vspace{-10pt}%
\caption{Total results of Llama2-7b-chat-hf, head ablation inference with log probability.
}
\label{fig:log_prop_app1}
\vspace{-10pt}
\end{figure*}

\begin{figure*}[t]
\vspace{-10pt}
\begin{center}
    \includegraphics[width=0.7\textwidth]{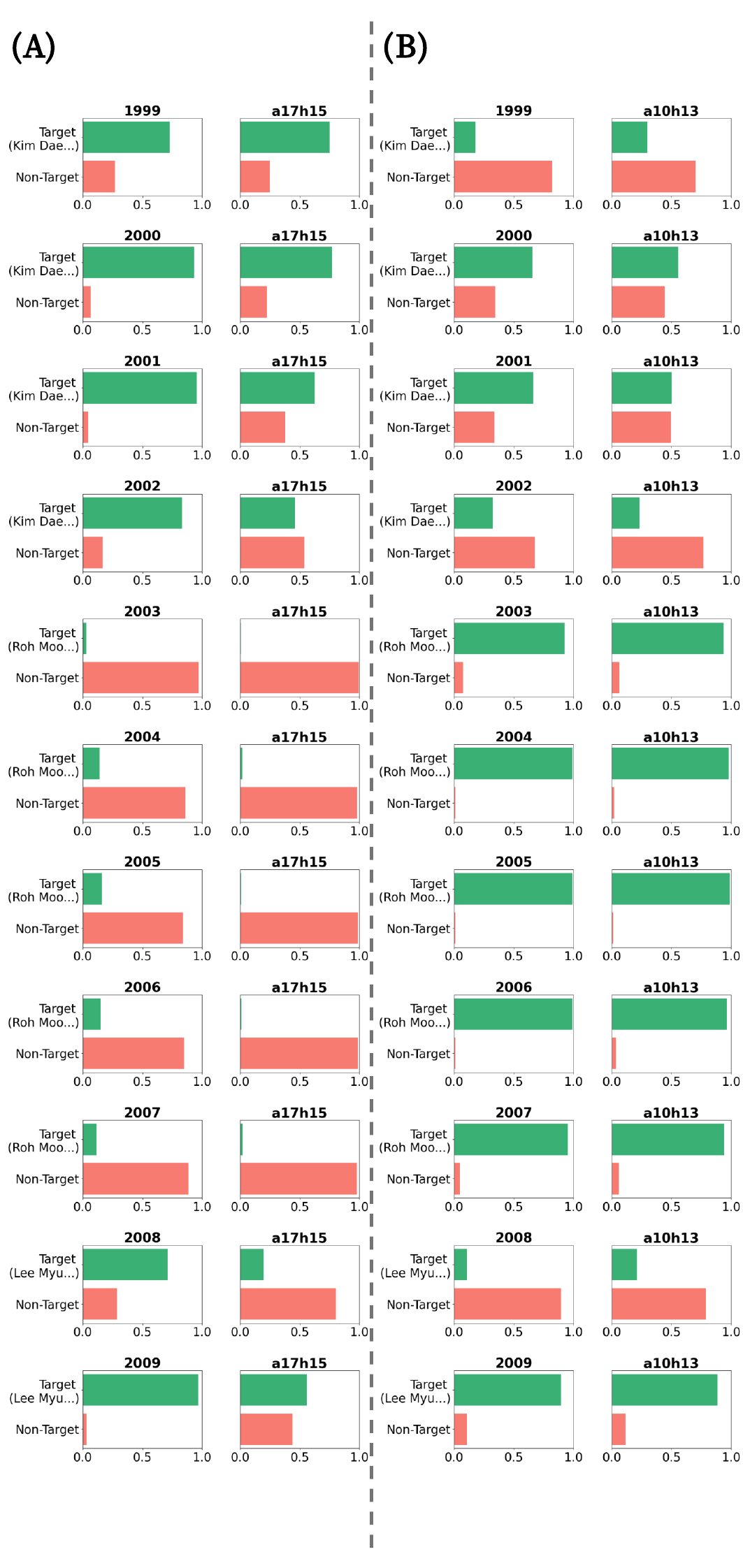}
\end{center}%
\vspace{-10pt}%
\caption{Total results of Qwen1.5-7B-Chat and Phi-3-mini-4k-instruct, head ablation inference with log probability.
(A) denotes the result of Qwen 1.5 and (B) represents the result of Phi 3 mini.
}
\label{fig:log_prop_app2}
\vspace{-10pt}
\end{figure*}

\begin{figure*}[t]
\vspace{-10pt}
\begin{center}
    \includegraphics[width=0.8\textwidth]{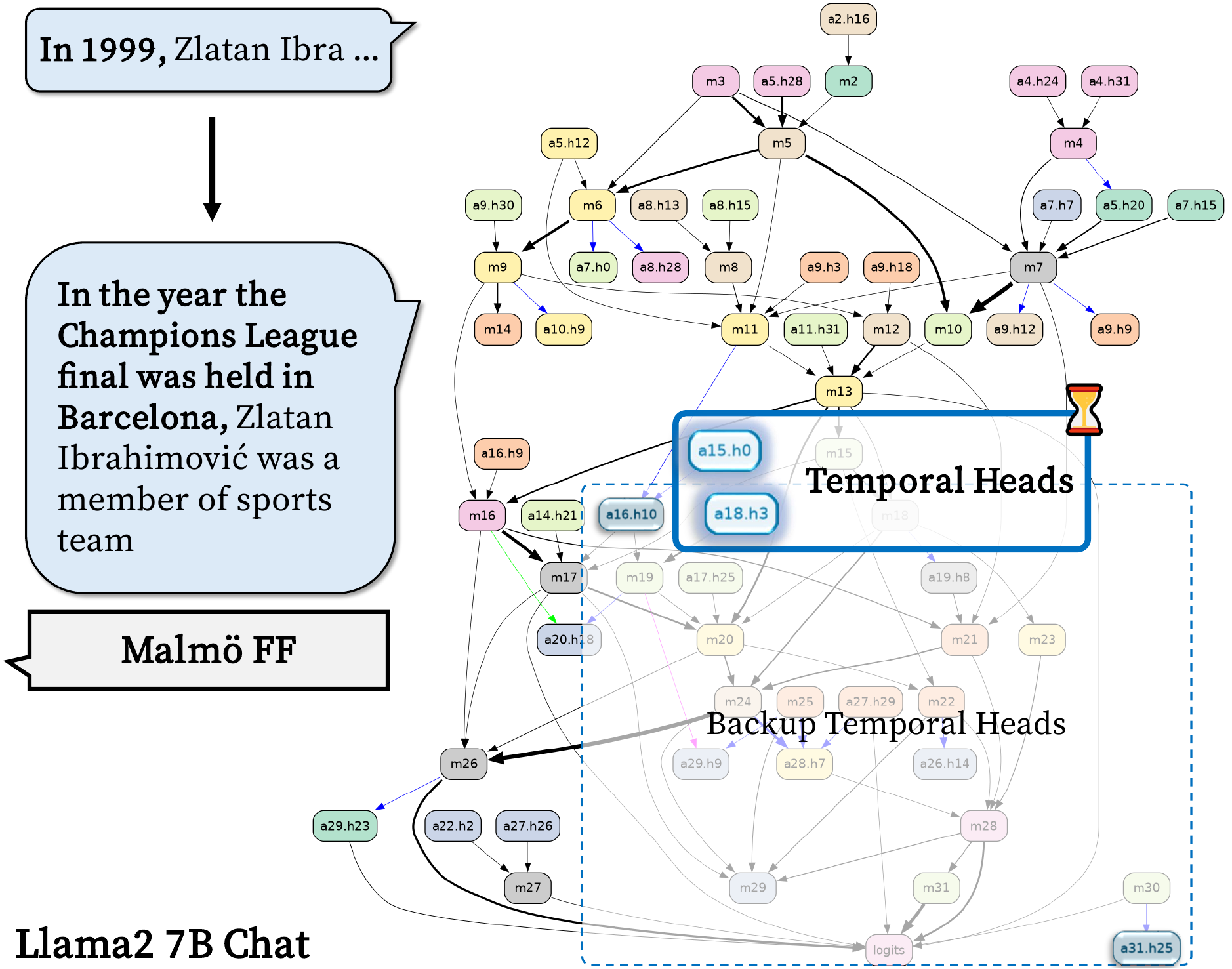}
\end{center}%
\vspace{20pt}
\begin{center}
    \includegraphics[width=0.8\textwidth]{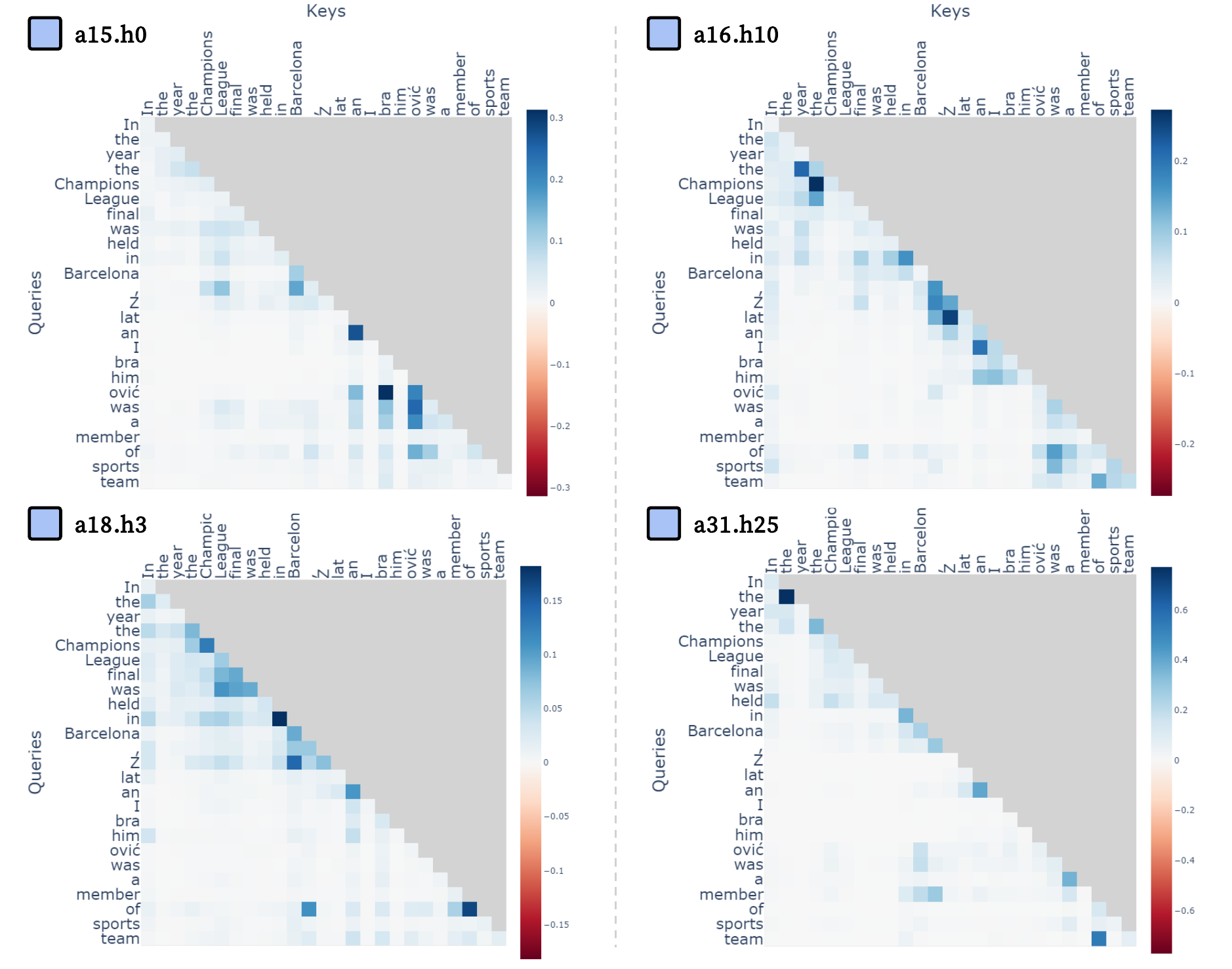}
\end{center}%
\caption{Temporal knowledge circuit from textual temporal conditioned prompt.
Here, we change the temporal condition \emph{"In 1999"} into \emph{"In the year the Champions League final was held in Barcelona"}, which model already correctly recalls the answer \textit{Malmö FF}.
The temporal knowledge circuit successfully catches temporal conditioning even with alias based on event based textual conditioning, with correctly showing off temporal knowledge heads and some backup temporal heads.
Figure of downside is the attention maps for each temporal heads and backup temporal heads.
Each of those figures highlight various tokens in conditioning part of prompt.
}
\label{fig:alias_app}
\vspace{-10pt}
\end{figure*}

\begin{figure*}[t]
\vspace{-10pt}
\begin{center}
    \includegraphics[width=0.8\textwidth]{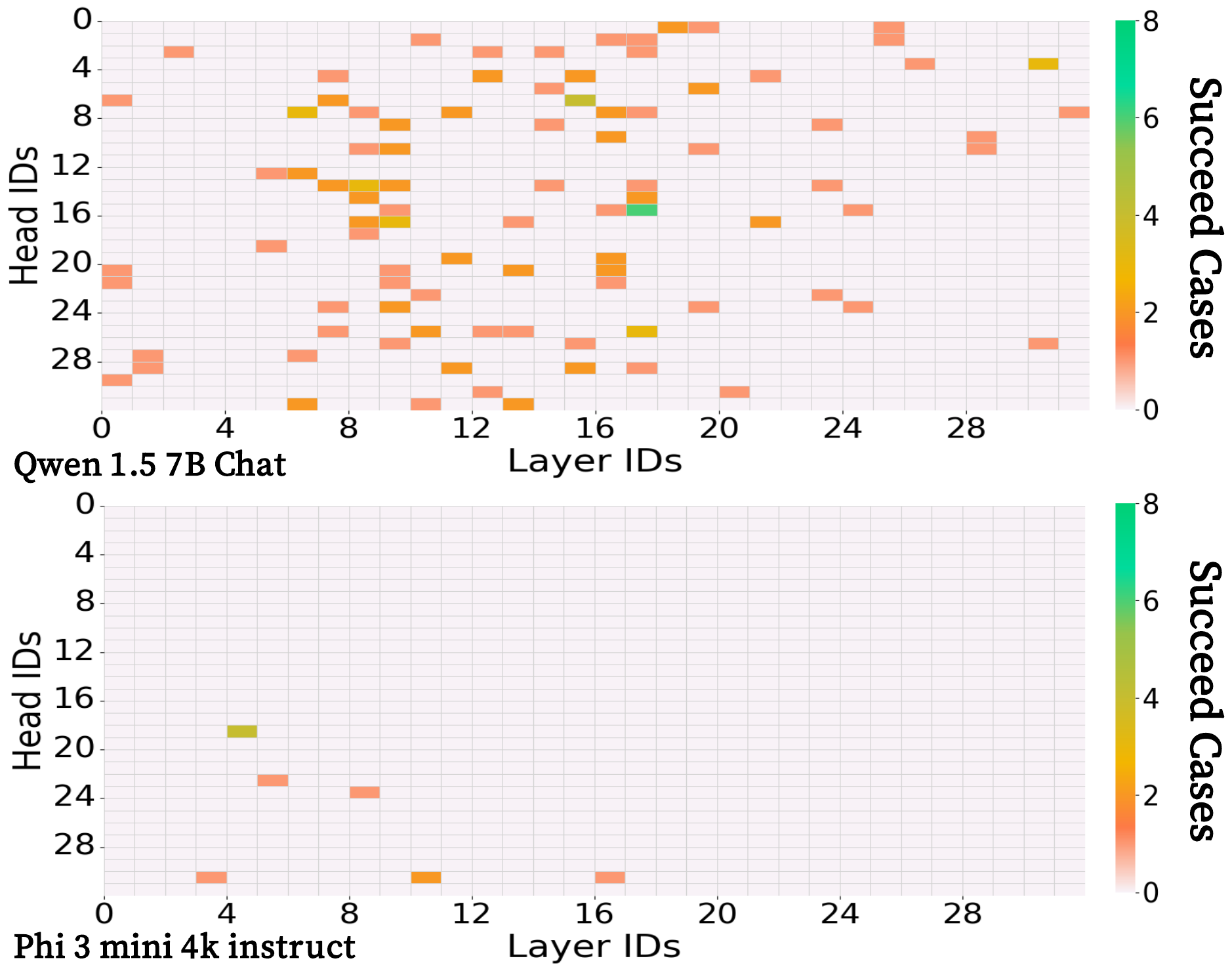}
\end{center}%
\caption{Result Of temporal knowledge editing in Qwen 1.5 7B Chat and Phi 3 mini 4k instruct.
From the source prompt, we catch the attention value of each model's temporal head, \textbf{a17.h15} and \textbf{a10.h13}.
The model's output is changed into temporally correct answer from temporally wrong answer.
The headmap below denotes the number of success in editing for every combination of layers and heads.
Though the most successful case of editing is the temporal head \textbf{a17.h15} in Qwen 1.5 7B Chat, Phi 3 mini 4k instruct shows that adding attention had minimal impact, and temporal heads failed to enable effective editing. 
This suggests that the model, constrained by its small parameter size (3.8B), requires a more sophisticated vector steering mechanism rather than relying on a single attention head value modification.
}
\label{fig:editing_app}
\vspace{-10pt}
\end{figure*}

\end{document}